\documentclass[lettersize,journal]{IEEEtran}
\usepackage{amsmath,amsfonts}
\usepackage{algorithm}
\usepackage{algorithmicx}
\usepackage{listings}
\usepackage{amssymb} 

\usepackage{array}
\usepackage{caption}

\usepackage{stfloats}
\usepackage{url}
\usepackage{verbatim}
\usepackage{comment}

\usepackage{graphicx}
\usepackage[font=small]{subcaption} 

\usepackage{cite}
\usepackage{longtable}
\usepackage{tabularray}
\usepackage{lscape}
\usepackage{xltabular}
\usepackage{multirow}
\usepackage{siunitx}
\usepackage{lscape}

\usepackage{listings}

\usepackage[normalem]{ulem}
\useunder{\uline}{\ul}{}
\usepackage{lscape}
\usepackage[usenames,dvipsnames,svgnames,table, x11names]{xcolor}
\usepackage[hyperfootnotes=false]{hyperref}
\hypersetup{
 colorlinks=true,
 citecolor=Blue,
 linkcolor=Red,
 urlcolor=Blue}

\usepackage{tikz}
\usetikzlibrary{positioning, fit, shapes, calc}
\usetikzlibrary{arrows.meta, arrows}
\usetikzlibrary{backgrounds} 

\usetikzlibrary{shapes.geometric}

\tikzstyle{startstop} = [rectangle, rounded corners,
minimum width=3cm, minimum height=0.7cm,
align=center, text width=1.8cm, inner sep=0.5mm, font=\small,
draw=black, 
fill=red!10]

\tikzstyle{storage} = [tape, 
minimum width=3cm, 
minimum height=1cm, 
align=center, 
text width=3cm, inner sep=0.5mm, font=\small,
draw=black, 
fill=orange!30]

\tikzstyle{io} = [trapezium, 
trapezium stretches=true, 
trapezium left angle=70, 
trapezium right angle=110, 
minimum width=3cm, 
minimum height=1cm, align=center, 
draw=black, fill=blue!10]

\tikzstyle{process-start} = [rectangle, 
minimum width=3cm, 
minimum height=0.8cm, 
align=center, 
text width=5cm, font=\small,
draw=black, 
fill=orange!10]

\tikzstyle{process-optional} = [rectangle, 
minimum width=3cm, 
minimum height=0.8cm, 
align=center, 
text width=5cm, font=\small,
fill=orange!10]

\tikzstyle{process} = [rectangle, 
minimum width=3cm, 
minimum height=1cm, 
align=center, 
text width=3cm, 
draw=black, 
fill=orange!30]

\tikzstyle{decision} = [diamond, 
minimum width=3cm, 
minimum height=1cm, 
align=center, 
draw=black, 
fill=green!30]
\tikzstyle{arrow} = [thick,->,>=stealth]

\tikzstyle{comments} = [rectangle, 
minimum width=10cm, 
minimum height=0.8cm, 
text width=10cm, font=\small,
draw=none, 
fill=orange!]

\newcommand{\seg}{%
  \begin{tikzpicture}[scale=0.37]
      
    \draw[thick] (0,0.5) -- (0.5,0);
    
    \fill[green!10] (0.25,0) -- (0.5,0) -- (0.5,0.25) -- (0.25,0.25) -- cycle;
    
    \fill[yellow!15] (0,0.25) -- (0.25,0.25) -- (0.25,0.5) -- (0,0.5) -- cycle;
    \fill[blue] (0,0) -- (0.3,0) -- (0.3,0.3) -- (0,0.3) -- cycle;
    \draw[] (0,0) rectangle (0.5,0.5);
  \end{tikzpicture}%
}

\newcommand{\OD}{%
  \begin{tikzpicture}[scale=0.19]
    \draw[] (0,0) rectangle (1,1);   

    \fill[black] (0.3,0.3) rectangle (0.7,0.7);
  \end{tikzpicture}%
}

\begin{document}

\title{Self-Supervised Learning for Image Segmentation: A Comprehensive Survey}

\author{~Thangarajah~Akilan$^{\dagger}$,~\IEEEmembership{Senior Member,~IEEE,} Nusrat~Jahan$^{\dagger}$,~\IEEEmembership{Graduate Student Member,~IEEE,}
Wandong~Zhang,~\IEEEmembership{Member,~IEEE}
\thanks{$^{\dagger}$contributed equally to this work.}
}

\markboth{Self-Supervised Learning for Image Segmentation: A Comprehensive Survey (PREPRINT)}%
{Shell \MakeLowercase{\textit{et al.}}: A Sample Article Using IEEEtran.cls for IEEE Journals}
\maketitle

\begin{abstract}
 
Supervised learning demands large amounts of precisely annotated data to achieve promising results. Such data curation is labor-intensive and imposes significant overhead regarding time and costs.
Self-supervised learning (SSL) partially overcomes these limitations by exploiting vast amounts of unlabeled data and creating surrogate (pretext or proxy) tasks to learn useful representations without manual labeling.
As a result, SSL has become a powerful machine learning (ML) paradigm for solving several practical downstream computer vision problems, such as classification, detection, and segmentation.
Image segmentation is the cornerstone of many high-level visual perception applications, including medical imaging, intelligent transportation, agriculture, and surveillance.
Although there is substantial research potential for developing advanced algorithms for SSL-based semantic segmentation, a comprehensive study of existing methodologies is essential to trace advances and guide emerging researchers.
This survey thoroughly investigates over 150 recent image segmentation articles, particularly focusing on SSL.
It provides a practical categorization of pretext tasks, downstream tasks, and commonly used benchmark datasets for image segmentation research.
It concludes with key observations distilled from a large body of literature and offers future directions to make this research field more accessible and comprehensible for readers.

\end{abstract}


\section{Introduction}

\begin{figure}[!tp]
     \includegraphics[trim={0cm, 0cm, 0cm, 0cm}, clip, width=1.0\columnwidth]{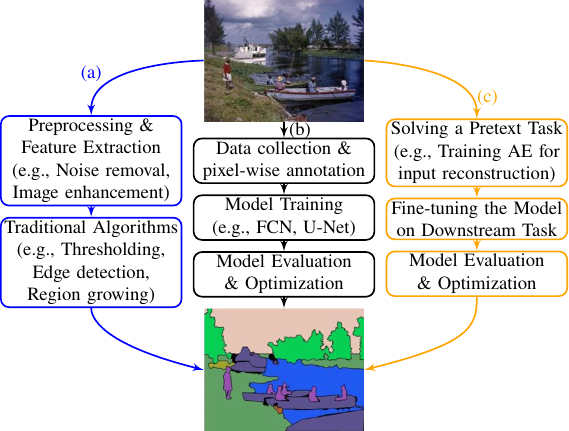}
    \footnotesize The pipelines \textcolor{blue}{(a)}, \textcolor{black}{(b)}, and \textcolor{Orange}{(c)}, represent traditional, supervised, and SSL approaches used for image segmentation,  respectively. Note: In (a), there is no explicit model evaluation since there is no training involved unlike in (b) or (c). The input and output images are adopted from \cite{kirillov2017unified}.    
    \caption{Three widely used image segmentation techniques.} \label{fig:seg-technique} 
\end{figure}

\IEEEPARstart{I}{mage} segmentation is a pixel-level classification process that partitions an input image into multiple segments that are more meaningful for high-level applications. For instance, segmentation generates foreground masks that delineate objects while suppressing background details, offering a more information-rich representation than simple bounding boxes. Such masks are particularly beneficial for applications like video surveillance.
Over the past few decades, various segmentation techniques have been developed in three main learning paradigms: traditional methods, end-to-end supervised deep learning (DL), and self-supervised learning, as illustrated in Fig.~\ref{fig:seg-technique}. Traditional approaches, including intensity thresholding, histogram-based clustering, edge detection, region growing, graph cuts, probabilistic models, conditional random fields~\cite{zhang2009image, kuccukkulahli2016histogram}, etc., rely on predefined heuristics, hypotheses, and preset hyperparameters rather than explicit training and model evaluation. These methods extract salient features from preprocessed input images before performing segmentation and can yield reliable results in controlled environments. 
However, their performance often deteriorates in real-world scenarios, where variations in input data are unpredictable.

Recent advancements in DL, have significantly enhanced image segmentation through end-to-end training on large-scale labeled datasets. DL architectures originally developed for image classification, 
have been effectively adapted for segmentation employing fully convolutional networks (FCNs), upsampling, transposed convolution (ConvT), and residual feature concatenation~\cite{9356353}. These innovations have gained state-of-the-art (SOTA) performance in medical image segmentation~\cite{mishra2018ultrasound}, moving object segmentation, and semantic segmentation~\cite{hao2020brief, araslanov2020single, Chen_2023_CVPR}.
Concurrently, semi-supervised and weakly-supervised learning approaches have received the attention of the computer vision community~\cite{novosel2019boosting, kalapos2023self, valada2020self, mahapatra2021interpretability, 10690208} due to their ability to reduce dependence on labeled data. This shift has driven a transition from traditional supervised learning to SSL.

\subsection{The Paradigm Shift from Supervised Learning to SSL}\label{sec-paradigm-shift}

SSL is a rapidly advancing field as it can learn latent features from unlabeled data. It employs pretext tasks, where pseudo-labels are generated directly from raw data, enabling model training without manual annotation. Once pre-trained, the model can be fine-tuned with minimal labeled data on a target domain~\cite{liu2025unsupervised, gui2024survey, you2024mine}. This process closely mirrors human learning, where knowledge is acquired incrementally through feedback and adaptation.
In contrast, supervised learning heavily depends on large annotated datasets, which are costly and time-intensive to create, particularly in complex domains such as medical image analysis. Moreover, supervised models are prone to generalization errors, spurious correlations, and adversarial attacks. SSL mitigates these issues by leveraging unlabeled data to capture intrinsic patterns, reducing dependence on annotations, and improving domain adaptation.
As a result, SSL has gained significant traction across various applications, including image classification~\cite{misra2020self, Wallin_2024_WACV, Caron_2024_WACV}, object recognition~\cite{zhai2019s4l, afouras2020self, azizi2021big, wang2023self}, image segmentation~\cite{sowrirajan2021moco, li2021rotation, xie2020pgl, Xu_2024_WACV}, and NLP applications, like emotion recognition~\cite{zhu2023joint, lan2019albert}.

\subsection{Existing Survey}\label{existing-survey} 

\begin{table*}[!ht]
\centering
\caption{\label{tab:existing} A comparative summary of existing surveys on self-supervised-driven techniques for visual data analysis.} 
\setlength{\tabcolsep}{2pt} 
\renewcommand{\arraystretch}{1.2} 
\begin{tabular}{
    >{\centering\arraybackslash}m{0.8cm} 
    >{\raggedright\arraybackslash}m{5cm} 
    >{\centering\arraybackslash}m{3cm} 
    >{\raggedright\arraybackslash}m{8.8cm}} 
\hline
\centering \textbf{Ref.} & \centering \textbf{Objective} & 
\centering \textbf{Targeted Application} & \multicolumn{1}{c}{\textbf{Contribution}} \\
\hline 
\hline

 \cite{jing2020self} 2021 & Categorization of pretext tasks   & \seg, \protect\OD, $\boxtimes$   & Covering general SSL pipelines, terminologies, deep neural networks, evaluation metrics, and datasets used for visual feature learning. \\ \hline

\cite{ohri2021review} 2021  & Comparative analysis on both supervised and SSL approaches  & \protect\OD, $\boxtimes$  & Highlighting recent works in SSL using contrastive learning and clustering methods, and listing of open challenges. \\ \hline

\cite{xu2021review}  2021 & Exploring SSL methods in the field of medical image analysis  & Medical image analysis & Categorizing the SSL techniques in medical image analysis, and comparing their performances on benchmark datasets. \\  \hline 
 
\cite{shurrab2022self} 2021  & Reviewing the SOTA methods in SSL for computer vision  & General computer vision \& medical image analysis  & A high-level overview of the state-of-the-art SSL methods in computer vision tasks and medical image analysis; grouping them as predictive, generative, and contrastive self-supervised methods. \\  \hline

\cite{albelwi2022survey} 2022  & A comprehensive overview of SSL using contrastive learning and auxiliary pretext & \seg, \protect\OD, $\boxtimes$ & Organizing various SSL techniques and examining recent research in this field; conducting performance analyses of different SSL approaches. \\  \hline

\cite{9356353} 2022  & Survey on DL-based image segmentation techniques & \seg & Exploring a wide range of cutting-edge semantic, and instance-level segmentation methods; investigating commonly used datasets. \\ \hline
\cite{gui2024survey} 2024 & Survey on SSL methods, covering their basics and application domains & Broad area of CV and NLP & Focused on visual SSL in four groups: context-based, contrastive learning, generative, and contrastive generative algorithms.  \\ \hline
\rowcolor{blue!5}
{This Survey} & Image segmentation-focussed a concise yet detailed investigation of SSL  & \protect\seg & A comprehensive overview of SSL approaches; an intuitive categorization of SSL-based image segmentation models with their application examples; a curated review of benchmark datasets; a depth analysis of SSL challenges in image segmentation and emerging research directions.  \\
\hline \hline 
\multicolumn{4}{l}{\footnotesize Note: \protect\seg~- Segmentation, \protect\OD~- Object detection, $\boxtimes$ - Classification, and SSL - self-supervised learning.}
\end{tabular} 
\end{table*} 

Table~\ref{tab:existing} summarizes existing key surveys, highlighting their objectives and contributions.
Chen~\textit{et~al.}~\cite{chen2020deep} review supervised DL methods for cardiac image segmentation, while Minaee~\textit{et~al.}~\cite{9356353} examine DL image segmentation models, analyzing their interconnections, strengths, and challenges. Liu~\textit{et~al.}~\cite{liu2023deep} explore DL approaches for brain tumor segmentation, discussing architectures, performance under imbalanced data, and multi-modality strategies.

However, few surveys focus on SSL, where most of them address SSL in classification \cite{jing2020self}, object detection \cite{huang2022survey}, image recognition \cite{ohri2021review}, and remote sensing \cite{wang2022self}. Xu~\cite{xu2021review} provides a concise overview of SSL frameworks, evaluating their performance on benchmark medical image datasets. 
Some studies examine SSL's domain adaptation for downstream applications. Rani~\textit{et~al.}~\cite{rani2023self} summarize such downstream tasks and relevant datasets. Hao~\textit{et~al.}~\cite{hao2020brief} group semantic segmentation models based on supervision levels, while Shurrab and Duwairi~\cite{shurrab2022self} survey SSL methods for medical imaging without analyzing model performance. Jing and Tian~\cite{jing2020self} classify SSL pretext tasks into four types and compare their performances.
Despite extensive research on image segmentation, most surveys emphasize supervised algorithms. Studies dedicated to self-supervised segmentation remain limited, underscoring the need for this work to bridge the gap.

The rest of this article is structured as follows: Section~\ref{terminology} lays a foundation of SSL. Section~\ref{ssl-img-seg} systematically reviews the SSL methods used for image segmentation. Section~\ref{dataset} lists benchmark datasets. Section~\ref{sec-future-challenges} discusses the challenges and directions for the prospect of SSL. Finally, Section~\ref{conclusion} concludes the article.

\subsection{Contributions}
Given SSL's effectiveness in addressing data scarcity, this article presents an in-depth comparative analysis of recent SSL-based segmentation algorithms. The key contributions of this study are: (i) A comprehensive review of self-supervised learning approaches, (ii) A systematic categorization of SSL-based image segmentation models and their applications, (iii) A curated review of benchmark datasets for training and evaluation, (iv) An analysis of SSL challenges in image segmentation and emerging research directions.

\section{Preliminary}\label{terminology}

\begin{figure*}[t]
\centering
\includegraphics[trim={0.2cm, 0.20cm, 0.6cm, 0cm}, clip, width=1\textwidth]{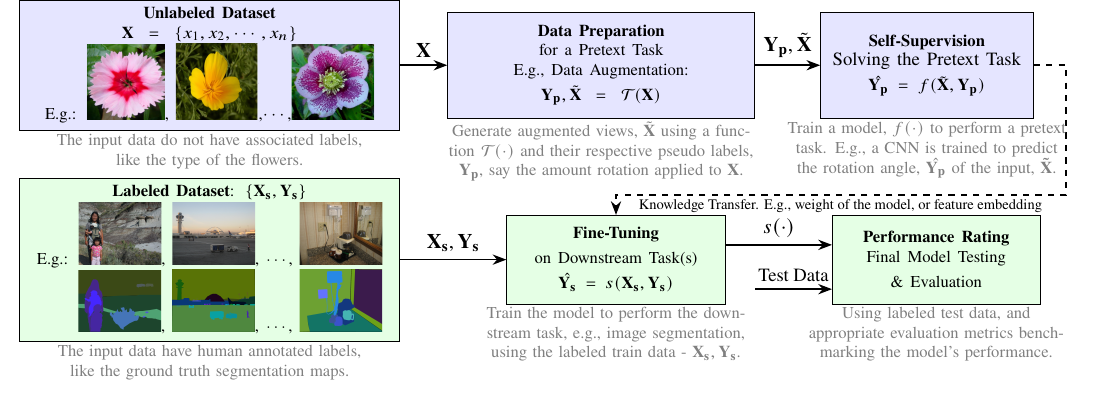}
    \caption{SSL-driven model development. (a) data preparation - $\mathcal{T}(\cdot)$ create synthetic samples $\mathbf{\tilde{X}}$ with pseudo labels $\mathbf{Y_p}$ from the unlabeled data $\mathbf{X}$ (b) self-supervised training - a model is trained using $\{\mathbf{\tilde{X}}, \mathbf{Y_p}\}$, (c) fine-tuning - the model is fine-tuned on a downstream task using a labeled dataset $\{\mathbf{X_s}, \mathbf{Y_s}\}$, and (d) performance rating - the final model is evaluated using test data of the target application. The unlabeled and labeled samples are adopted, respectively from \cite{Nilsback06} and \cite{zhou2017scene}.}
    \label{fig:work-process}
\end{figure*}

\subsection{Self-supervised Learning} 

Fig.~\ref{fig:work-process} on page \pageref{fig:work-process} illustrates the general workflow of SSL-driven model building. 
It involves (i) defining pretext label-free task, (ii) data preparation for the pretext task, (iii) representation learning via self-supervision, (iv) knowledge transfer, (v) fine-tuning on downstream task, and (vi) performance rating. 

\subsubsection{Pretext Task}\label{sec-pretext}

It serves as the cornerstone of SSL, enabling self-supervision without the need for manually annotated labels. The core idea is to start with simpler pretext tasks and progressively refine the model for more complex downstream tasks.
Various pretext tasks, viz. jigsaw puzzles~\cite{freeman1964apictorial}, rotation prediction~\cite{gidaris2018unsupervised}, context encoding~\cite{li2025anatomask, pathak2016context}, sequence prediction~\cite{georgescu2021anomaly, 10.1007/978, mohamed2022self}, and image denoising~\cite{xie2020noise2same, kazimi2024self}—are designed to extract domain-specific features~\cite{albelwi2022survey, xu2019self}. These tasks facilitate learning generalized representations from raw data, which are then exploited for downstream applications.
Pretext and downstream tasks often share semantic overlap, allowing multiple pretext tasks to be combined for robust feature learning~\cite{10531186, shi2023video}. However, designing an effective pretext task requires careful formulation: (i) \textbf{balancing complexity}—tasks that are too simple may fail to capture meaningful features, while overly complex tasks risk overfitting or learning irrelevant patterns; (ii) \textbf{domain alignment}—ensuring the pretext task is structured to prevent shortcut learning and effectively transfer knowledge to the downstream objective.

\subsubsection{Representation Learning via Self-supervision}
It involves extracting useful, generalizable features from raw data, as the model optimizes an objective function while solving a pretext task. Each pretext task is tailored to the input data type and target application, leading to variations in the objective function.
The goal is to guide models
to learn meaningful representations using synthetically generated inputs from the unlabeled data, $\tilde{\mathbf{X}}$, and their pseudo-labels, $\mathbf{Y_p}$. A generic objective function is defined as the average loss over $N$ samples in a mini-batch:
\begin{equation}\label{eq:loss-D} 
    \mathcal{L} = \frac{1}{N} \sum_{n=1}^{N} J(\tilde{X}^n, Y_p^n),
\end{equation}
where $\tilde{X}^n$ and $Y_p^n$ are the $n$th input sample and its pseudo-label, respectively, and $J(\cdot)$ computes the per-sample loss (the difference between the predicted label $\hat{Y_p}$ and pseudo-label $Y_p$).
The effectiveness of SSL hinges on the quality of the representations that encode essential input features, thereby improving generalization across diverse downstream tasks.

\subsubsection{Fine-tuning on Downstream Task}\label{sec-downstream}

A downstream task is the model’s intended application~\cite{10690208, kotar2021contrasting}. In this phase, the pre-trained representation learning model is adapted using labeled data of the target domain, with fine-tuning involving partial or full retraining and integration of a task-specific top layer.
During fine-tuning, the target task-specific objective functions and evaluation metrics are applied to optimize and assess model performance.

\begin{figure}[!tp]
    \includegraphics[trim={0cm, 0cm, 0cm, 0cm}, clip, width=1\columnwidth]{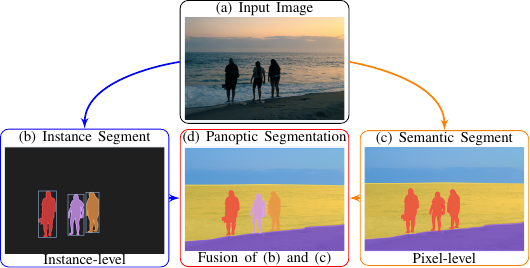}
    \scriptsize {The input and output images are adopted from \href{https://www.v7labs.com/blog/panoptic-segmentation-guide}{https://www.v7labs.com}}.
    \caption{For a given (a) Input, (b) instance segmentation: per-object mask and class label, (c) semantic segmentation, and (d) panoptic segmentation: per-pixel class + instance-level labels.} \label{fig:panoptic-seg} 
\end{figure}

\subsection{Image Segmentation}

Image segmentation assigns pixel-level labels to partition an image into sub-regions, serving as a key pillar for detailed scene understanding and various applications.
Segmentation approaches fall into three main categories as illustrated in Fig.~\ref{fig:panoptic-seg}: instance segmentation~\cite{kirillov2019panoptic, li2017fully, yang2023review}, semantic segmentation~\cite{10458980, 10462518, brostow2009semantic}, and panoptic segmentation~\cite{yin2024revisiting, 10380457, hu2023you}, which unifies the first two. Among these, semantic segmentation has seen widespread adoption in domains, viz. agriculture, healthcare, infrastructure management, and transportation~\cite{marsocci2021mare}.

For example, in healthcare, deep convolutional neural networks (DCNNs) are used for tasks, like tumor segmentation from brain MRIs~\cite{chen2019multi, zhang2023self} and whole-heart segmentation from computed tomography (CT) images~\cite{zhang2021self}. 
Similarly, precision agriculture relies on segmentation for insect detection, leaf disease identification, flower and crop recognition, farmland anomaly analysis, etc. 
However, applying deep supervised learning to image segmentation still faces 4 major challenges:
(i) handling high-resolution inputs,
(ii) accommodating diverse object shapes and sizes,
(iii) requiring large-scale, densely annotated datasets, and
(iv) achieving accurate delineation of segmentation boundaries~\cite{7913730, patel2023multi, zhang2021self, NIPS2014_07563a3f}.
Recent works address these challenges using label-efficient SSL~\cite{guldenring2021self, zhao2023cla, chen2021exploring, siddique2022self}.

\section{Pretext Task for Image Segmentation}\label{ssl-img-seg}

\begin{figure*}[!htp]
\centering
\includegraphics[trim={0.8cm, 0.1cm, 0.5cm, 0.0cm}, clip, width=1\textwidth]{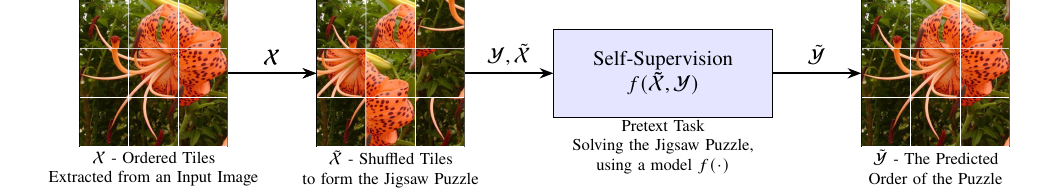}
    \caption{Illustration of the Jigsaw puzzle. An input image is divided into titles ($\mathbf{\mathcal{X}}$) and then shuffled ($\tilde{\mathbf{\mathcal{X}}}$), where $\mathbf{\mathcal{Y}}$ is the pseudo label denoting the correct order of the titles. The model is trained to rearrange the shuffled titles to the correct order. 
    }  
    \label{fig:jigsaw}
\end{figure*}

To perform image segmentation, various types of pretext tasks are employed in SSL. These pretext tasks can be broadly categorized into three groups based on the nature of the learning objective: predictive methods, generative methods, and contrastive methods~\cite{novosel2019boosting, jing2020self, Nikita2021CVF, li2021semanticJnl, ye2024cads}.

\subsection{Predictive Methods}

Predictive methods generate pseudo-labels by applying transformations or masking strategies to infer specific properties of the input data. Common predictive pretext tasks include slice order prediction, jigsaw puzzles, Rubik’s cube problems, and rotation prediction. However, the effectiveness of these tasks varies, as their utility depends on their alignment with the underlying objectives of the segmentation task.

\subsubsection{Jigsaw puzzle}\label{sec-jigsaw} 

Jigsaw puzzle-solving is a classical pattern recognition problem involving the reassembly of disordered visual fragments, first addressed by Freeman and Garder in 1964~\cite{freeman1964apictorial}. In self-supervised image segmentation, it serves as a pretext task to train a model, $f(\cdot)$, to learn spatial relationships between image patches, as shown in Fig.~\ref{fig:jigsaw}.  It involves partitioning an image into patches, shuffling them, and training the model to recover the original arrangement. By solving this spatial reasoning problem, the model learns useful representations for downstream tasks.

Noroozi and Favaro~\cite{noroozi2016unsupervised} introduced this approach by extracting nine non-overlapping $80\times80$ patches from an image, labeling each by its original position, and training a model to predict the correct order when given a shuffled sequence. The trained model functions as a feature extractor (backbone), with a downstream task-specific sub-network fine-tuned on the target dataset.  
Yang~\textit{et~al.}~\cite{yang2022fully} extended this method by increasing the number of patches to 25, introducing slight overlaps, and enlarging the patch size to $115\times115$. These modifications improved semantic segmentation performance by capturing finer low-level features, such as edge and texture continuity, essential for precise segmentation.  
Beyond natural image segmentation tasks, the jigsaw puzzle has also been used in medical image analysis. 
For example, Taleb~\textit{et~al.}~\cite{taleb2021multimodal} applied this for CT scan and MRI image segmentation tasks. They considered the puzzle as a self-supervised reconstruction problem, in which a model is trained by minimizing the reconstruction error between the sorted ground truth patch sequence, $P^*$, and the reconstructed version of the shuffled input patch sequence, $P$ as in \eqref{eq-jigsaw-mse}.
\begin{equation}
 \mathcal{L}_{\text{puzzle}}(\theta, P, P^*) = \sum_{i=1}^{K} \left\| P_i^* - S_{\theta, P_i}^T\cdot P_i \right\|^2, \label{eq-jigsaw-mse}
\end{equation}
where $\theta$ and $K$ stand for the model parameters, and the total number of training puzzles, respectively. Hence, $S_{\theta, P_i}^T\cdot P_i$ is the reconstructed version of the i-th puzzle. 
Through training, the model would capture different tissue structures that were important for the downstream task. Thus, Taleb~\textit{et~al.} fine-tuned the model in various target domains for segmentation (prostate segmentation, brain tumor segmentation, liver segmentation) and regression problems (patient survival days prediction). 

Building on earlier works, this puzzle-solving has been further explored as a pretext task for image segmentation~\cite{ni2024drcl}, clustering~\cite{song2023grid}, video anomaly detection~\cite{wang2022video}, medical image classification~\cite{park2024fine}, and out-of-distribution detection~\cite{yu2024exploring}.  
Advanced variants, such as 3D jigsaw and spatiotemporal jigsaw, introduce additional complexity, enabling models to learn more robust spatial, temporal, and geometric relationships, thus improving downstream performance~\cite{xu2021review, markaki2023jigsaw}.  
However, this approach has key \textbf{limitations}: model performance is highly sensitive to puzzle tile size and permutation complexity, and it incurs substantial computational costs.

\subsubsection{Slice Order Prediction} 

\begin{figure}[!tp]
    \includegraphics[trim={0.9cm, 0.0cm, 0.2cm, 0.0cm}, clip, width=1\columnwidth]{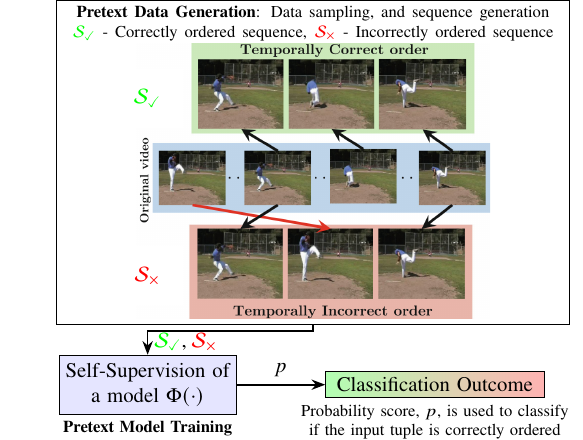}
    \caption{Illustration of slice order prediction pretext task. The image inside the block is adopted from \cite{misra2016shuffle}.}  
    \label{fig:order_predict_pretext}
\end{figure}

Ordering, a basic concept in mathematics and combinatorial optimization, predates modern computing. Early mechanical devices, such as Babbage's analytical engine~\cite{bromley1982charles}, laid the groundwork for contemporary sorting algorithms.  
In computer vision, ordering is framed as a sequential verification task, akin to the jigsaw puzzle~\cite{ zhukov2020learning, misra2016shuffle, zhang2017self, shvetsova2023learning, yang2024made}, but primarily applied to volumetric data (e.g., MRI, CT, PET). It captures the inherent 3D structure by learning spatial relationships between 2D slices.  
As illustrated in Fig.~\ref{fig:order_predict_pretext} on page \pageref{fig:order_predict_pretext}, the task is formulated as: (i) \textbf{pretext data generation}—sampling a data volume or video into 2D frames, generating correctly ordered ($\mathcal{S}_\checkmark$) and incorrectly ordered ($\mathcal{S}_\times$) sequences; and (ii) \textbf{self-supervised training}—training a model, $\Phi(\cdot)$, to maximize probability $p$ of correctly ordered tuples. 
Alternatively, some studies formulate this as a reconstruction problem~\cite{10.1007/978, yang2024made, jayaraman2016slow, srivastava2015unsupervised}.  
Formally, given a set of images, $\mathcal{I} \in \mathbb{R}^{C \times H \times W}$, the model reconstruct an attribute map, $\mathbf{S}_{\text{rec}} \in \mathbb{R}^{F \times H \times W}$ and sequence order $\mathbf{y}_{\text{order}} \in \mathbb{Z}^F: y_{\text{order}, i} \in \{0, 1, 2, \dots, F-1\}$ as in \eqref{eq-slice-order}.

\begin{equation}
    \mathbf{S}_{\text{rec}} = \Phi(\mathcal{I}_0, \mathcal{I}_1, \dots, \mathcal{I}_{F-1}), \label{eq-slice-order}
\end{equation}
where the model $\Phi$ is trained to order an arbitrary sequence of $F$ images shuffled from a volumetric data or temporal sequence.
For example, in \cite{10.1007/978}, a multi-task representation learning strategy is introduced with two pretext tasks aimed at predicting (i) the next slice and (ii) the previous slice in a sequence from a randomly selected slice of a volumetric CT scans. 
This strategy helps the model learn to map from $x^j_i$ to $x^{j-t}_i$ and $x^{j+t}_i$, capturing the structural similarity, including the relative location, shape, size, and structure of the organs. The model is optimized using the standard Euclidean distance defined by \eqref{eq-l2-mtl}, which serves as the objective function.
\begin{equation}
   \mathcal{L}_{pretext} = \sum_i \Vert(f_\theta^-(x^j_i)-x^{j-t}_i)\Vert + \Vert (f_\theta^+(x^j_i)-x^{j+t}_i)\Vert, \label{eq-l2-mtl}
\end{equation}
where $f_\theta^-(\cdot)$, and $f_\theta^+(\cdot)$ represent the models, in this case, decoder sub-networks that perform the backward sequence prediction and forward sequence prediction, respectively. While, $x^j_i$, $x^{j-t}_i$ and $x^{j+t}_i$, denote the $j$th, $(j-t)$th, and $(j+t)$th slices, respectively, in the $i$th CT scan. 
Similarly, Spitzer~\textit{et~al.}~\cite{spitzer2018improving} developed slice ordering as a geodesic distance prediction task along the brain surface, to learn spatial relationships between adjacent slices in 3D medical images. 
This task can be extended to anomaly detection, where frame sequences are shuffled, and the model predicts the correct temporal order or reconstructs the original video~\cite{misra2016shuffle, wang2020order, yang2024made}.  
 
\begin{figure}[!tp]
\centering
     \includegraphics[trim={0cm, 0cm, 1.2cm, 0cm}, clip, width=1.0\columnwidth]{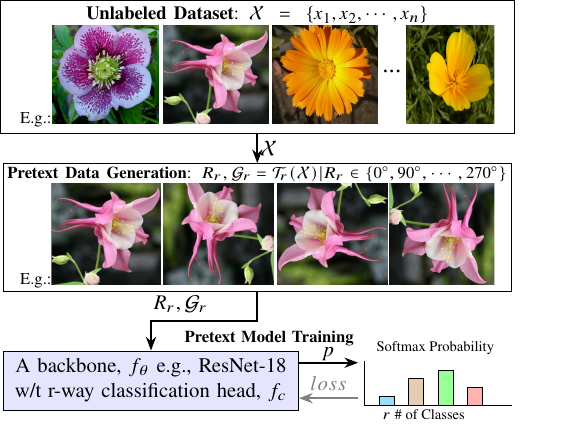}
        \caption{Overview of the rotation prediction pretext task. 
        }
    \label{fig:rotation}
\end{figure}

\subsubsection{Rotation Prediction} 
In this task, a geometric transformation, specifically rotation, is applied to the input image, and a model is trained to predict the applied transformation through classification or regression~\cite{gidaris2018unsupervised}. Prior research has demonstrated that rotation prediction can be an effective self-supervised pretext task for downstream applications, such as object detection, recognition, and segmentation.  
For instance, \cite{feng2019self} and \cite{deng2021does} employ a transformation function $\mathcal{T}_r(\cdot)$, which rotates an input image $x_1$ by an angle $R_r$ randomly chosen from a predefined set $\{0^\circ, 90^\circ, 180^\circ, 270^\circ\}$, generating the transformed sample $\mathcal{G}_r$, as illustrated in Fig.~\ref{fig:rotation}. 
A model $f_\theta(\cdot)$, parameterized by $\theta$, is then trained to predict the rotation angle using a classification head $f_c$ by minimizing the cross-entropy loss ($\mathcal{L}_{\text{CE}}$) between the true one-hot encoded label $R_r$ and the predicted probability distribution $f_c(f_\theta(\mathcal{G}_r))$:
\begin{equation}
    \mathcal{L}_{\text{rot}} = \frac{1}{4} \sum_{r \in \{0^\circ, 90^\circ, 180^\circ, 270^\circ\}} \mathcal{L}_{\text{CE}}(R_r, f_c(f_\theta(\mathcal{G}_r))), \label{eq-rotation}
\end{equation}
where $\mathcal{L}_{\text{rot}}$ denotes the mean classification loss over four rotations, and $f_c(\cdot)$ outputs a Softmax probability distribution over rotation classes.  

In addition to conventional classification tasks, \cite{deng2021does} explores co-training self-supervised rotation prediction alongside supervised tasks, assessing its impact on robustness and generalization under distribution shifts. Their findings indicate that self-supervised pretext tasks can improve model resilience in challenging environments. Meanwhile, \cite{feng2019self} mitigates rotation ambiguity by designing a nonparametric similarity-based approach that distinguishes instance-specific and rotation-invariant features, yielding superior performance across classification, detection, and segmentation tasks. 
Additionally, Linmans~\textit{et~al.}~\cite{linmans2018sample} integrate rotation-equivariant CNN architectures with rotation prediction pre-training to improve tumor segmentation.

Despite its advantages, rotation prediction exhibits certain \textbf{limitations}. Its effectiveness varies across architectures and tasks, as the pre-learned features may not generalize optimally. Moreover, the reliance on discrete rotation angles limits the model’s ability to capture continuous transformations, potentially making learned representations overly task-specific and less adaptable to real-world variations.  

\subsubsection{Rubik's Cube Recovery} 

\begin{figure*}[!tp]
\centering
\includegraphics[trim={1.4cm, 0cm, 0.3cm, 0.7cm}, clip, width=0.89\textwidth]{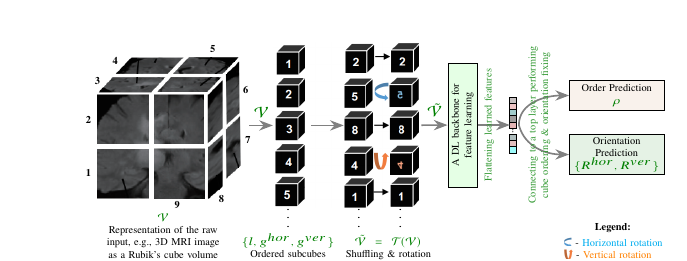}
    \caption{Illustration of the Rubik’s cube recovery-like pretext task. The input is divided into subcubes and a problem is formulated by cube shuffling, then a model is trained to solve the problem. The figure is recreated based on \cite{zhuang2019self}. 
    }  
    \label{fig:rubikcube}
\end{figure*}

This task combines the Jigsaw puzzle~\cite{noroozi2016unsupervised}, slice order prediction~\cite{10.1007/978}, and rotation prediction~\cite{gidaris2018unsupervised} tasks. As shown in Fig.~\ref{fig:rubikcube}, it involves two operations: cube rearrangement and application of rotation, enabling the model to learn translational and rotational invariant features from volumetric data, such as 3D medical images~\cite{zhuang2019self, zhu2020rubik, tao2020revisiting}.
A given sample $\mathcal{V}$ with dimensions $W \times H \times L$ is divided into subcubes of size $W/n \times H/n \times L/n$, where $n$ is the side length of each subcube. The initial configuration of these subcubes, determined by their order $l$ and orientations $g^{\text{hor}}$ and $g^{\text{ver}}$ in the horizontal and vertical directions, respectively, represents the solved state (ground truth) of the Rubik’s cube.
A series of random volume-wise transformations, $\mathcal{T}(\mathcal{V})$, creates a Rubik's cube puzzle, $\mathcal{\tilde{V}}$, with the solved state serving as the ground truth for self-supervised training. The model is optimized by minimizing the objective function:
\begin{equation}
   \mathcal{L}_{rubik} = \gamma\mathcal{L}_p + \eta\mathcal{L}_r, \label{eq-rubik}
\end{equation}
where $\mathcal{L}_p$ and $\mathcal{L}_r$ are the order permutation and orientation loss, respectively, and $\gamma$ and $\eta$ control the relative weight of each term.
When $K$ rotations applied, and the model predicts $\rho$ with one-hot label $l$, the order permutation loss is:
\begin{equation}
   \mathcal{L}_{p} = -\sum_{i=1}^K l_i \log(\rho_i), \label{eq-permutation}
\end{equation}
where $l_i$ is the true label and $\rho_i$ is the predicted probability for $i$th transformation.
The orientation prediction is treated as a multi-class classification, where the model predicts the rotation status 
of each subcube. Let $N$ be the number of subcubes, and $g$ be the ground truth vector with $1$ for rotated cubes and $0$ otherwise. The model outputs two $N$-dimensional vectors, $R^{hor}$ and $R^{ver}$, representing horizontal and vertical rotation probabilities, respectively. The orientation loss is defined as:
\begin{equation}
   \mathcal{L}_r = -\sum_{n=1}^N \left[g_n^{hor} \log(R_n^{hor}) + g_n^{ver} \log(R_n^{ver})\right], \label{eq-orientation}
\end{equation}
where $g_n^{hor}$ and $g_n^{ver}$ are the ground truth values for horizontal and vertical rotations, and $R_n^{hor}$ and $R_n^{ver}$ are the predicted probabilities.

Zhuang~\textit{et~al.}~\cite{zhuang2019self} focus on hemorrhage classification and brain tumor segmentation in 3D medical images, where they use Rubik’s cube recovery task to self-supervise a CNN. For classification, the CNN is fine-tuned directly on the target dataset. For segmentation, the pre-trained weights are adapted to the encoder, with feature maps expanded using dense upsampling Conv to generate pixel-wise predictions, replacing the ConvT used in standard encoder-decoder models. This reduces the trainable parameters and mitigates random initialization effects. Their results show improvements of $2.3\%$ in brain tumor segmentation.

As an extension, \cite{zhu2020rubik} and \cite{tao2020revisiting} enhance this task by adding random cube masking. 
This modification leads to the updated objective function: \begin{equation} 
\mathcal{L}_{rubik} = \gamma\mathcal{L}_p + \eta\mathcal{L}_r + \delta\mathcal{L}_m, \label{eq-rubik+} \end{equation} 
where $\mathcal{L}_p$, $\mathcal{L}_r$, and $\mathcal{L}_m$ represent the order permutation, orientation, and mask identification losses, respectively. The mask identification loss is a multi-classification problem: \begin{equation}
   \mathcal{L}_{m} = -\sum_{n=1}^N g^m_n \times \log(\rho^m_n), \label{eq-mask-rubik}
\end{equation} 
with $g_n^m$ indicating masked cube position and $\rho_n^m$ being the predicted probability, for the $n$th subcube. This addition challenges the model to learn translation, rotation, and noise-invariant features.
Unlike previous works, \cite{tao2020revisiting} employs a GAN-based architecture to combine Rubik's cube reconstruction and adversarial losses, improving SSL for 3D medical image segmentation. This approach increases the results of brain tissue segmentation in the MRBrainS18~\cite{E0U32Q_2024} dataset by $3.97\%$ in the dice score compared to training from scratch.

\textbf{Limitations:} While fine-tuning Rubik’s cube recovery-based pre-trained models outperforms training-from-scratch models in tasks, like brain hemorrhage classification and tumor segmentation, this approach primarily captures spatio-global representations. Consequently, it may lack the fine-grained pixel-level details required for precise segmentation boundaries. Moreover, incorporating complexities, such as subcube masking or adversarial learning can further increase the computational overhead of the self-supervision pipeline.

\subsection{Generative Methods}

\begin{figure*}[!tp]
\centering
    \includegraphics[trim={0.1cm, 0.1cm, 0cm, 0.0cm}, clip, width=1\textwidth]{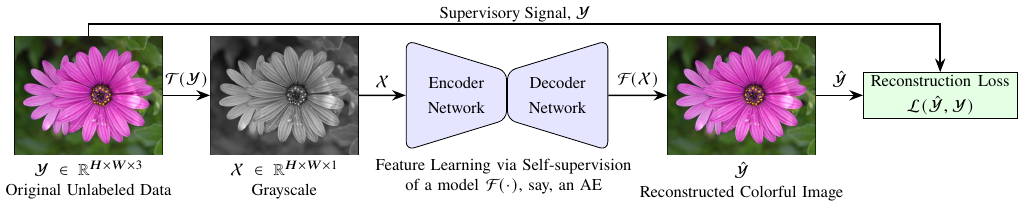}
    \caption{Encoder-decoder network predicts the probable color from a grayscale image. $\mathcal{T}(\cdot)$ is an RGB to grayscale transformation.}  
    \label{fig:colorization} \vspace{-0.2cm}
\end{figure*}

In this pretext task, the model reconstructs the input, predicts missing parts (e.g., masked images or cloze tests in NLP~\cite{liu2021self}), or generates new data. These tasks range from low-level detail reconstruction to high-level semantic modeling. The methods train an encoder to map inputs to latent feature representations and a decoder to reconstruct them~\cite{soliman2020s, zeng2019sese, felfeliyan2023self}.
Advanced techniques, including adversarial learning, perceptual losses, and context encoders, are effective in capturing semantic information. However, simpler methods, like early GANs or AEs often struggle with long-range semantic relationships due to limited receptive fields or lack of contextual awareness.
Recent innovations, viz. larger receptive fields, dilated convolutions, diffusion models~\cite{croitoru2023diffusion}, and attention mechanisms~\cite{li2023ms}, improve the model’s ability to capture global context and semantic coherence~\cite{treneska2022gan, vzeger2021grayscale}.

\subsubsection{Image Colorization} 

It is a fundamental generative pretext task in SSL-based image segmentation~\cite{gonzalez2021self, chen2024color}. As shown in Fig.~\ref{fig:colorization}, it trains a model to generate a colorized image from a grayscale input, leveraging contextual cues, such as edges and object boundaries\cite{zhang2016colorful}.
Given a grayscale input image, $\mathcal{X} \in \mathbb{R}^{H \times W \times 1}$, the objective is to learn a mapping $\hat{\mathcal{Y}} = \mathcal{F}(\mathcal{X})$ to the respective color image $\mathcal{Y} \in \mathbb{R}^{H \times W \times 3}$, where $H$, $W$ are spatial dimensions of the input. 
If we consider the representation of the RGB image as a 3-dimensional space, then a natural objective function is the Euclidean loss $\mathcal{L}_2(\cdot, \cdot)$ between predicted and ground truth colors defined in \eqref{eq-eucludian-colorization}. 
\begin{equation}\label{eq-eucludian-colorization}
    \mathcal{L}_2(\hat{\mathcal{Y}}, \mathcal{Y}) = \frac{1}{2} \sum_{h, w} \|\mathcal{Y}_{h,w} - \hat{\mathcal{Y}}_{h,w}\|_2^2.
\end{equation}

Although widely used in regression tasks, Euclidean loss in RGB space fails to capture multimodal color distributions due to the non-uniformity of RGB color perception. Perceptually uniform color spaces, such as CIE Lab~\cite{zhang2016colorful}, better align with human vision. Thus, multimodal color distributions are often modeled as a multinomial classification task over a quantized color set~\cite{larsson2016learning, zhang2016colorful}. For a given input \( \mathbf{X} \), the model learns a mapping \( \mathcal{Z} = \mathcal{G}(\mathbf{X}) \), where \( \mathcal{Z} \in [0,1]^{H \times W \times Q} \) is the probability distribution over the \( Q \) quantized color at each pixel location \((h, w)\), satisfying \( \sum_q \mathcal{Z}_{h, w, q} = 1 \). The training objective is typically a multinomial cross-entropy loss: 
\begin{equation}
    \mathcal{L}_{cl}(\hat{\mathcal{Z}}, \mathcal{Z}) = -\sum_{h, w} v(\mathcal{Z}_{h, w}) \sum_{q} \mathcal{Z}_{h, w, q} \log(\hat{\mathcal{Z}}_{h, w, q}), \label{eq-mn-ce}
\end{equation}
where \( \hat{\mathcal{Z}} \) and \( \mathcal{Z} \) denote the predicted and ground-truth color distributions, respectively. The term \( v(\cdot) \) is a reweighting factor that adjusts the loss based on the rarity of specific color classes to mitigate the imbalance of classes. 
This framework can also be adapted for color correction tasks. For example, \cite{LIN2023109021} presents a self-supervised GAN framework with CIE Lab color channel correction to learn light-invariant features for plant leaf segmentation. While colorization and color correction do not directly involve segmentation, they enable the model to distinguish regions and objects, aiding in contextual understanding. As a result, the learned features are effectively transferred to semantic segmentation tasks~\cite{ho2025every, chen2024color, LIN2023109021}.

\subsubsection{Image De-noising} 

\begin{figure}[!tp]
\centering
     \includegraphics[trim={0cm, 0cm, 0cm, .1cm}, clip, width=0.99\columnwidth]{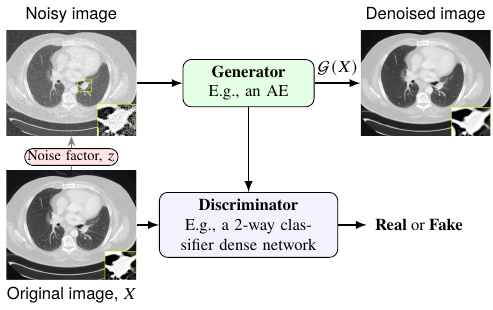}\\\vspace{-1cm}
     {\hspace{2.8cm} \scriptsize  Note: The CT scan sample images are adopted from \cite{kim2020image}} \vspace{0.1cm}
     \caption{Conceptual diagram of a GAN-based denoising.}\label{fig:denoising-method} \vspace{-0.2cm}
\end{figure}

\begin{figure*}[!tp]
\centering
    \includegraphics[trim={0.1cm, 0.1cm, 0.0cm, 0.1cm},clip, width=.99\textwidth]{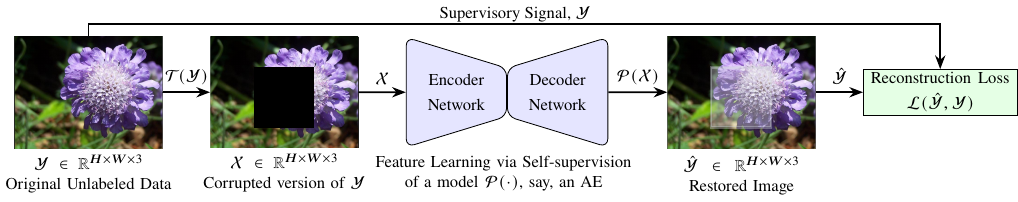}
    \caption{The concept of the self-supervised image inpainting pre-training process using an AE. }
    \label{fig:context-encoder} \vspace{-0.2cm}
\end{figure*}

It is typically tackled using models, like GANs similar to those used for image colorization~\cite{xie2020noise2same, kazimi2024self}. 
GANs, pioneered by \cite{goodfellow2014generative}, were extended by Luc~\textit{et~al.}~\cite{luc2016semantic} to semantic segmentation and further developed by Hung~~\textit{et~al.}~\cite{hung2018adversarial}, Belagali~~\textit{et~al.}~\cite{belagali2024gen}, and Li~~\textit{et~al.}~\cite{li2021semantic}, for semi-supervised segmentation. Fig.~\ref{fig:denoising-method} illustrates a GAN-based framework, consisting of two modules: a generator $\mathcal{G}$, which maps noisy inputs ${X}$ to denoised outputs $\mathcal{G}({X})$, and a discriminator $D$, which evaluates whether a sample belongs to the real data distribution.

The framework is trained using the adversarial loss function $\mathcal{L}_{GAN}$:  
\begin{equation}\label{eq:LcGAN} 
\mathcal{L}_{GAN}(G, D) = \mathbb{E}_{x}[\log D(x)] + \mathbb{E}_{x}[\log(1 - D(\mathcal{G}(x)))],
\end{equation}
where $D(x)$ is the discriminator's output for real samples, and $D(\mathcal{G}(x))$ is its output for the generator's denoised prediction $\mathcal{G}(x)$. The generator learns to map noisy inputs to clean images, minimizing the discrepancy from the real data distribution $\mathbb{E}_{x}[\log(1 - D(\mathcal{G}(x)))]$, while the discriminator aims to distinguish between real and generated images by maximizing the term $\mathbb{E}_{x}[\log D(x)]$, the log probability of correctly classifying real samples. This adversarial optimization ensures the generator progressively improves its denoising capability.  
The generator can then be fine-tuned on domain-specific datasets. For example, \cite{kazimi2024self} fine-tune it with limited labeled data for semantic segmentation, achieving competitive results.  
Beyond GANs, AEs are widely used for image denoising. Denoising autoencoders (DAEs) learn to reconstruct clean images from corrupted inputs. Their training objective, $\mathcal{L}(\theta)$, is typically defined as the mean squared error (MSE) between the input and its reconstructed output:  
\begin{equation}
    \mathcal{L}(\theta) = \frac{1}{N} \sum_{i=1}^N ( f_\theta(x_i + \epsilon_i) - x_i )^2, \label{eq:de_noise}
\end{equation}
where $ f_\theta(\cdot) $ is the DAE function, $ x_i $ is the input image, $ \epsilon_i $ is added noise, and $ N $ is the number of training samples.  

This approach has been successfully applied in microscopy data segmentation. For instance, \cite{prakash2020leveraging, martinez2021impartial} use DAEs for nuclei segmentation, demonstrating that denoising pretext tasks improves segmentation accuracy. DAEs can also be trained with masked inputs, focusing loss computation on unmasked pixels, which enhances object boundary reconstruction.  
In medical imaging, Zhang~\textit{et~al.}~\cite{zhang2021self} propose a 3D-AE for self-supervised tumor segmentation in MRI. They reconstruct normal organs, tumors, and segmentation masks separately, using L1 loss for organ/tumor reconstruction and Dice loss for segmentation mask prediction. The overall objective is:  
\begin{equation}
    \mathcal{J} = \lambda_1\mathcal{L}_{org} + \lambda_2\mathcal{L}_{tum} + \lambda_3\mathcal{L}_{dic},
\end{equation}
where $\mathcal{L}_{org}$ and $\mathcal{L}_{tum}$ are L1 reconstruction losses for normal and tumor regions, and $\mathcal{L}_{dic}$ is the Dice loss for segmentation masks, with $\lambda$'s controlling the contribution of each term. They also incorporate transformations, elastic deformations, and Gaussian blurring to regulate the difficulty of self-supervision.  
Overall, both GANs and AEs are effective in SSL-based segmentation, as they learn robust representations while mitigating noise sensitivity. They also provide an efficient way to pre-train deep models on large-scale unlabeled data.

\subsubsection{Image Inpainting} 

In this task, a model reconstructs missing regions of a corrupted image by predicting masked patches~\cite{ zhang2023optimized, he2022semantic, katircioglu2020self, pathak2016context, caron2024location}, as illustrated in Fig.~\ref{fig:context-encoder}. This pretext task enables the model to learn intrinsic representations, capturing structured, semantically meaningful characteristics that improve contextual understanding and output quality.
Referring to Fig.~\ref{fig:context-encoder}, the original image $\mathcal{Y}$ undergoes corruption through a transformation function $\mathcal{T}(\cdot)$, defined as:
\begin{equation} 
\mathcal{X} = \mathcal{T}(\mathcal{Y}) = M \odot \mathcal{Y}, \label{eq-mask} 
\end{equation} 
where $\odot$ denotes element-wise multiplication, and $M \in \{0,1\}^{H \times W}$ is a binary mask. Pixels with $M = 0$ are removed, while those with $M = 1$ are retained.
In traditional inpainting, a model $\mathcal{P}$ restores the original image by minimizing the normalized masked $L_2$ reconstruction loss:
\begin{equation} 
    \mathcal{L}_{\text{rec}}(\mathcal{X}) = \frac{1}{\sum (1 - M)} \| (1 - M) \odot (\mathcal{Y} - \mathcal{P}(\mathcal{X})) \|_2^2, \label{eq-inpainting-recon-loss} 
\end{equation} 
where $\sum (1 - M)$ normalizes the loss by the number of corrupted
pixels. Research~\cite{pathak2016context} suggests that $L_1$ and $L_2$ losses yield comparable performance, with selection depending on data characteristics.
The study~\cite{he2022semantic} showed that this pretext task enhances semantic segmentation of remote sensing images, particularly when fine-tuning with limited labeled data. Similarly,~\cite{katircioglu2020self} demonstrated that self-supervised learning with Monte Carlo optimization improves human detection and segmentation.
GAN-based approaches~\cite{pan2020unsupervised} have also been explored for inpainting. Performance can be further improved by modifying the masking function, integrating context-aware encoding, and employing attention mechanisms~\cite{wu2024image}. 


\subsubsection{Context Restoration} 

\begin{algorithm}[!tp]
\caption{Image context swapping algorithm}
\label{algo-context-swap} 
\begin{lstlisting} 
def image_context_disordering(x_i, T):
    # Input: x_i - Input image (array-like or tensor)
    #        T - Number of iterations (int)
    # Output: x_tilde_i - Img w/t disordered context
    
    x_tilde_i = x_i.copy()
    
    for iter in range(1, T + 1):
        # Randomly select two non-overlapping patches from x_tilde_i
        p1 = randomly_select_patch(x_tilde_i)
        p2 = randomly_select_patch(x_tilde_i)

        # Ensure the patches do not overlap
        if not overlap(p1, p2): 
            # Swap the selected patches
            swap(p1, p2)
    return x_tilde_i
\end{lstlisting} \vspace{-0.3cm}
\end{algorithm}

It is built upon image inpainting and slice-order prediction techniques, involving two key subtasks: (i) identifying corrupted regions and (ii) reconstructing the correct image context. 
Instead of masking random regions, this method swaps the contexts of two non-overlapping regions, $p_1$ and $p_2$ ($p_1 \cap p_2 = \emptyset$), for $T$ iterations, as detailed in Algorithm~\ref{algo-context-swap}.
After the patch-swapping, the corrupted image is denoted as $\tilde{X}$. The model then learns the parameters $\theta$ by minimizing an appropriate loss function, like MSE (cf.~\eqref{eq:de_noise}), to restore the original image ($\hat{X} = \mathcal{F}_\theta(\tilde{X})$). 
 For instance, Chen~\textit{et~al.}~\cite{chen2019self} applied this pretext task with $64 \times 64$ patches to enhance classification, segmentation, and localization in medical imaging modalities, viz. brain MRI, abdominal CT, and ultrasound spine coronal images.
However, this approach has the following \textbf{limitations}: (i) selecting an appropriate patch size is crucial, as small patches may fail to capture strong contextual features, reducing the pretext task's effectiveness; (ii) determining the optimal number of iterations is challenging, as excessive corruption can hinder the model’s ability to restore context, rendering the task counterproductive. 
Hence, pixel-wise generative SSL is computationally expensive and struggles to learn abstract latent patterns~\cite{xie2020pgl}.

\subsection{Contrastive Methods}

\begin{figure}[!tp]
\centering
    \includegraphics[trim={0cm, 0cm, 0cm, 0.22cm}, clip, width=0.95\columnwidth]{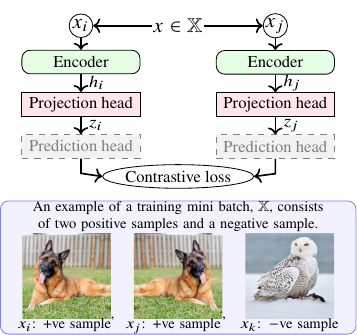}
    \caption{The concept of self-supervised contrastive learning. }    \label{fig:contrastive-learning}
    \vspace{-0.2cm}
\end{figure}

Fig.~\ref{fig:contrastive-learning} illustrates a general framework for contrastive learning, where a model is trained by minimizing the distance between positive ($+$ve) sample pairs, $\{x^+, x\}$, and maximizing the distance between negative ($-$ve)  sample pairs,  $\{x^-, x\}$, using the noise contrastive estimation (NCE) loss \cite{li2025anatomask, liu2021self, 9460820}:  
\begin{equation}
\text{NCE}_{x, x^+, x^-} = - \log \left( 
\frac{e^{f(x)^\top f(x^+)}}{e^{f(x)^\top f(x^+)} + e^{f(x)^\top f(x^-)}} 
\right),
\end{equation}
where  $f(\cdot) $ is a vision encoder (e.g., ViT),  $x^+$ is a similar sample to  $x$, and  $x^-$ is a dissimilar sample. The NCE loss maximizes the softmax probability of the positive pair while minimizing it for the negative pair.

The selection of positive and negative samples plays a critical role. Positive pairs, $\{x_i, x_j\}\in \mathbb{X}$ are usually generated using random data augmentations 
from the same input. Because these two pairs originate from the same source, their features are expected to be close to each other in the feature spaces, $h$ and $z$. 
Negative samples are the augmented views derived from different input samples in the same batch $\mathbb{X}$. 
Both augmented views $x_i$ and $x_j$ pass through a shared encoder (e.g., a DL backbone) to ensure consistency and alignment. The encoder maps the inputs to high-dimensional feature representations $h_i$ and $h_j$, which are then projected into a lower-dimensional latent space via a projection head, typically a shallow neural network (e.g., an MLP), yielding $z_i$ and $z_j$. 
The prediction head is optional. Some frameworks, such as bootstrap your own latent (BYOL)~\cite{grill2020bootstrap}, introduce a prediction head to infer one view’s representation from the other. This approach is a defining characteristic of non-contrastive learning methods, where negative samples are not required. However, prediction heads can also be incorporated into contrastive frameworks. 
Finally, a contrastive loss is applied in the latent space, pulling $+$ve pairs $\{z_i, z_j\}$ closer, while pushing $-$ve pairs $\{z_i, z_k\}$ or $\{z_j, z_k\}$ apart, where $z_k$ represents a $-$ve sample, to learn discriminative and invariant representations of the input.
The standard NCE loss can be generalized to information noise-contrastive estimation (InfoNCE), which maximizes a lower bound on mutual information by using multiple $-$ve samples ($x_k^-$) to improve discriminative power:
\begin{equation}
\text{\small InfoNCE}_{x, x^+, x_k^-} = 
- \log \left(\frac{e^{f(x)^\top f(x^+)}}{e^{f(x)^\top f(x^+)} + \sum_{k=1}^{K} e^{f(x)^\top f(x_k^-)}}\right). \label{InfoNCE}
\end{equation}

While specific objective functions and feature extraction models may vary, the fundamental framework remains consistent. Several contrastive learning methods, including Deep InfoMax (DIM) \cite{hjelm2019learning}, SimCLR \cite{chen2020simple}, and MoCo \cite{he2020momentum}, have been developed to enhance feature representation learning, achieving SOTA results across diverse domains. For instance, DIM maximizes mutual information between global and local features, while SimCLR and MoCo employ InfoNCE-based objectives to contrast positive and negative samples.
Thus, contrastive learning has proven effective in leveraging synthetic or limited labeled data \cite{she2021contrastive, karimijafarbigloo2023self}, demonstrating its potential for generalizable image-level representation learning.

\subsubsection{Contrastive Predictive Coding (CPC)} 

It was inspired by the InfoNCE loss and originally developed for learning representations from sequential data, like speech and text~\cite{oord2018representation, liu2021self}. It has since been extended to images to capture spatial information~\cite{henaff2020data}.
In CPC, an image is divided into overlapping patches $x_{i,j}$, where $(i,j)$ denotes the spatial location. Each patch is encoded using a shared model, $f_\theta$ (e.g., ResNet), to produce feature maps $z_{i,j} = f_\theta(x_{i,j})$. These feature maps are processed by another model, $g_\phi$ (e.g., a masked CNN), to compute context vectors $c_{i,j}$, which depend only on past spatial locations $(u \leq i, v \leq j)$, ensuring causality: $c_{i,j} = g_\phi(z_{u,v} | u \leq i, v \leq j)$.
The goal is to predict `future' feature maps $z_{i+k,j}$ from $c_{i,j}$ using a linear transformation with an embedding matrix $W_k$: $\hat{z}_{i+k,j} = W_k \cdot c_{i,j}$. A contrastive loss \eqref{eq-cpc} enforces correct recognition of the target $z_{i+k,j}$ among randomly sampled negative feature maps $z_l$~\cite{henaff2020data}:
\begin{equation}
    \mathcal{L}_{\text{CPC}} = - \sum_{i,j,k} \log p(z_{i+k,j} | \hat{z}_{i+k,j}, \{z_l\}), \label{eq-cpc}
\end{equation}
where \[ p(z_{i+k,j} | \hat{z}_{i+k,j}, \{z_l\})=\frac{\exp(\hat{z}_{i+k,j}^\top z_{i+k,j})}{\exp(\hat{z}_{i+k,j}^\top z_{i+k,j})+\sum_l \exp(\hat{z}_{i+k,j}^\top z_l)}.\]

Extensions, such as~\cite{wang2021dense}, and  \cite{hu2021region} adapt CPC for semantic segmentation by learning dense, localized representations for spatial patches instead of global representations. These modifications enhance pixel-wise feature learning, improving segmentation performance on benchmarks, like Cityscapes~\cite{cordts2016cityscapes}, ADE20K~\cite{xie2021selfsupervised}, COCO Stuff~\cite{caesar2018cvpr}, and PASCAL VOC~\cite{pascal-voc-2012}.
For 3D data, conventional CPC shows poor results due to the fundamental differences between 2D images and 3D volumes, particularly in spatial dependencies and feature extraction~\cite{zhu2020embedding}. To address this limitation, Zhu~\textit{et al.}~\cite{zhu2020embedding} propose task-related CPC (TCPC), which integrates task-specific knowledge to enhance the representation learning. 
TCPC modifies the InfoNCE loss to differentiate positive samples (same context or structure) from negative ones. 
It employs supervoxel estimation to locate lesions and extracts features from sub-volumes around these regions, using a calibrated contrastive scheme for self-supervised learning. Experimental results show that integrating clinical priors, such as lesion positions, improves performance in 3D medical classification tasks (e.g., brain hemorrhage and lung cancer detection using CT scans), outperforming proxy tasks like 3D Jigsaw puzzles and Rubik's cube solving.

However, CPC and its variants have a few \textbf{limitations}:\quad\quad (i) It has difficulty capturing long-range dependencies, making it less effective for tasks requiring holistic scene understanding (e.g., object detection in cluttered images); (ii) By focusing on local patch, it may overfit to low-level features (e.g., textures) rather than learning high-level semantic representations transferable across tasks; (iii) Processing overlapping patches incurs high computational costs, exacerbated by contrastive loss calculations over multiple pair of samples.

\begin{figure}[!tp]
\centering
    \includegraphics[trim={0.2cm, 0cm, 0.2cm, 0.1cm}, clip, width=1.0\columnwidth]{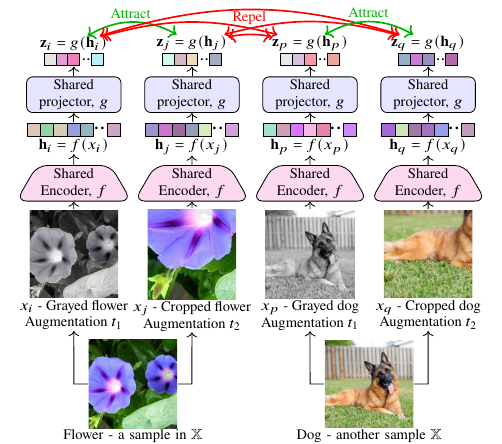}\\ 
    {\scriptsize Note: This figure is recreated based on \href{https://research.google/blog/advancing-self-supervised-and-semi-supervised-learning-with-simclr/}{Chen (2020)'s illustration of SimCLR}~\cite{chen2020simple}} 
    \caption{A SimCLR pipeline. The encoder and projector networks are trained together by minimizing a contrastive loss.
    }
    \label{fig:SimCLR}\vspace{-0.2cm}
\end{figure}

\subsubsection{Simple Framework for Contrastive Learning (SimCLR)} 

SimCLR simplifies the CPC pipeline by removing sequence prediction and focusing on instance discrimination. It learns invariant representations by maximizing agreement between two augmented views of the same image
\cite{chen2020simple}  as illustrated in Fig.~\ref{fig:SimCLR}.

During training, a minibatch $\mathbb{X}$ of $N$ images is sampled, and two random augmentations, $\mathcal{T} = \{t_1, t_2\}$, are applied to each image ($\tilde{x}_i = t_1(x)$, $\tilde{x}_j = t_2(x)$), generating $2N$ augmented samples. 
Given a positive pair $\{\tilde{x}_i, \tilde{x}_j\}$, the remaining $2N-2$ samples act as negatives. 
A shared encoder, $f(\cdot)$, extracts features: $\mathbf{h}_i = f(\tilde{x}_i)$ and $\mathbf{h}_j = f(\tilde{x}_j)$. These are processed by a projection head, $g(\cdot)$, to obtain embeddings: $\mathbf{z}_i = g(\mathbf{h}_i)$ and $\mathbf{z}_j = g(\mathbf{h}_j)$. Training is guided by the normalized temperature-scaled cross-entropy loss (NT-Xent):
\begin{equation}
    \mathcal{L}_{\text{SimCLR}}^{(i,j)} = -\log \frac{\exp(\text{sim}(\mathbf{z}_i, \mathbf{z}_j) / \tau)}{\sum_{k=1}^{2N} \mathbf{1}_{[k \neq i]} \exp(\text{sim}(\mathbf{z}_i, \mathbf{z}_k) / \tau)}, \label{eq-simclr-loss}
\end{equation}
where $\mathbf{1}_{[k \neq i]}$ ensures that the denominator excludes the similarity of $\mathbf{z}_i$ itself, i.e., 1 if $k \neq i$, 0 otherwise, $\text{sim}(\cdot)$ is cosine similarity, and $\tau$ is a temperature parameter. After training, the projection head can be discarded to meet the needs of downstream tasks.
Although originally designed for classification, recent adaptations enable SimCLR for image segmentation~\cite{gandhi2024self, torpey2024deepset, tang2023semantic, landgraf2022evaluation}, demonstrating its broader applicability.
A key \textbf{limitation} of SimCLR is its dependence on large batch sizes to ensure sufficient negative samples, raising computational concerns. Hence, it is sensitive to the quality and diversity of applied augmentations.

\subsubsection{Momentum Contrast (MoCo)} 

\begin{figure}[!tp]
\centering  
    \includegraphics[trim={5.2cm, 0.3cm, 8.6cm, 0.8cm}, clip, width=1\columnwidth]{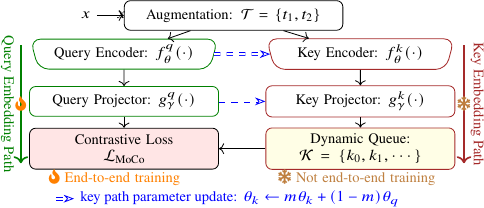} 
    \caption{An illustration of the MoCo pipeline. 
    }
    \label{fig:moco} \vspace{-0.2cm}
\end{figure}

MoCo, like SimCLR, is an instance-based contrastive learning approach but estimates contrastive loss using one positive pair and $K$ negative samples~\cite{he2020momentum}. It maintains a dynamic queue as a memory bank, storing features (keys) from previous training iterations beyond the current mini-batch. The queue maintains a fixed size of $K$ (e.g., 65,536) samples by removing the oldest keys on the fly.
Unlike SimCLR, MoCo does not use shared networks for both views. Instead, it maintains separate networks for query and key embeddings, as illustrated in Fig.~\ref{fig:moco}. Its training is asymmetric, i.e., only the query path, $f_\theta^q(\cdot)$ + $g_\gamma^q(\cdot)$, is updated via backpropagation using a modified \texttt{InfoNCE}, while the key embedding path is updated via a momentum update with smoothing exponentially moving average (EMA). For instance, the key encoder parameter, $\theta^k$, is updated as:
\begin{equation}
    \theta_k \leftarrow m\theta_k + (1-m)\theta_q, \label{eq-moco}
\end{equation}
where $m \in [0,1)$ is a momentum coefficient, and $\theta_q$ is the query encoder parameter. The same update rule applies to the key projection network, $g_\gamma^k(\cdot)$.
MoCo optimizes contrastive loss using the dynamic queue:
\begin{equation}
    \mathcal{L}_{\text{MoCo}} = -\log \frac{\exp(\text{sim}(\mathbf{q}, \mathbf{k}^+)/\tau)}{\sum_{i=1}^{K} \exp(\text{sim}(\mathbf{q}, \mathbf{k}_i)/\tau)}, \label{moco-InfoNCE}
\end{equation}
where $\mathbf{q}$ is the query representation, $\mathbf{k}^+$ is the positive key (an augmented view), $\mathbf{k}_i$ are negative keys (from the queue), $\tau$ is a temperature parameter, and $\text{sim}(\cdot, \cdot)$ denotes cosine similarity.

MoCo and its variants, including MoCo v2~\cite{chen2020improved}, MoCo v3~\cite{chen2021empirical}, and Fast-MoCo~\cite{ci2022fast}, have demonstrated effectiveness in semantic segmentation. 
These methods outperform fully supervised training across data types, viz. natural images, medical imaging, and Lidar point clouds~\cite{wang2024automated, xu2024swin, mahmoudi2024cloudspam, ci2022fast, chen2021empirical, chen2020improved, he2020momentum}.

MoCo's \textbf{limitations} are: (i) The dynamic queue increases memory usage and computational overhead, making it resource-intensive; (ii) Maintaining separate query and key networks adds computational and implementation complexity compared to SimCLR’s shared network; (iii) Sensitivity to the momentum coefficient $m$ poses challenges—high $m$ slows adaptation, while low $m$ risks instability.

\subsubsection{Bootstrap Your Own Latent (BYOL)} 

\begin{figure}[!tp]
\centering  
    \includegraphics[trim={13.7cm, 0.3cm, 0.1cm, 0.8cm}, clip, width=1\columnwidth]{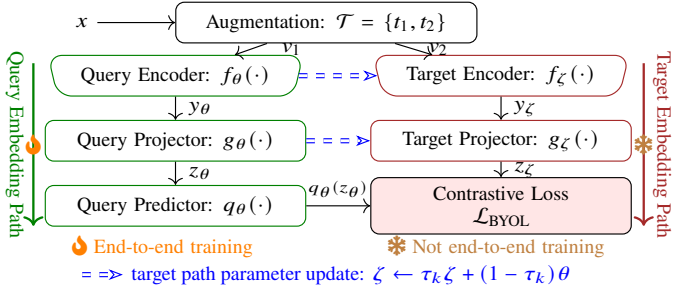}
    \caption{An illustration of the BYOL pipeline. 
    {Note: This figure is inspired by Grill (2020) \cite{grill2020bootstrap}.}}
    \label{fig:byol}\vspace{-0.2cm}
\end{figure}

While traditional contrastive learning methods (e.g., SimCLR, or MoCo) rely on distinguishing $+$ve and $-$ve pairs, BYOL adopts a self-distillation approach that learns without $-$ve samples by using only $+$ve pairs~\cite{grill2020bootstrap}. This is achieved through an additional prediction head, $q_\theta(\cdot)$, on top of the query projection head, as shown in Fig.~\ref{fig:byol}. The prediction head aligns query and target embeddings while preventing representation collapse.
The target network mirrors the query path, consisting of an encoder $f_\zeta(\cdot)$ and a projector $g_\zeta(\cdot)$, but with separate parameters $\zeta$. 
The query network (encoder, projector, and predictor) is updated via backpropagation, while the target network is updated solely through EMA of the query parameters. At training step $k$, for a momentum decay rate $\tau_k \in [0, 1)$, the update rule is:
\begin{equation}
    \zeta \leftarrow \tau_k \zeta + (1 - \tau_k) \theta. \label{eq-byol-target-param}
\end{equation}

During training, an image $x$ is transformed into two augmented views, $v_1$ and $v_2$, using augmentation $\{t_1, t_2\} \sim \mathcal{T}$. The query path computes the encoder features $y_\theta = f_\theta(v_1)$, projection map $z_\theta = g_\theta(y_\theta)$, and prediction  $q_\theta(z_\theta)$. The target path processes $v_2$ to produce $y_\zeta = f_\zeta(v_2)$ and projection $z_\zeta = g_\zeta(y_\zeta)$. The loss function is the mean squared error between the $\ell_2$-normalized prediction and target projection:
\begin{equation}
    \mathcal{L}_{BYOL} = 2 - 2 \cdot \frac{\langle q_\theta(z_\theta), z_\zeta \rangle}{\|q_\theta(z_\theta)\|_2 \cdot \|z_\zeta\|_2}, \label{eq:byol-loss}
\end{equation}
where $\langle \cdot, \cdot \rangle$ denotes cosine similarity. Minimizing $\mathcal{L}_{BYOL}$ enhances alignment between query and target projections, with zero loss indicating perfect alignment.

Initially proposed for image classification, BYOL has been extended to image segmentation~\cite{liu20243d, nisar2024maximising, grill2020bootstrap}, object detection~\cite{gao2024self}, and monocular depth estimation~\cite{grill2020bootstrap}.
However, BYOL has the following \textbf{limitations}: (i) Like MoCo, its dual-network design increases computational and memory overhead; (ii) Relying solely on positive samples may limit its ability to learn discriminative features, particularly in diverse or complex datasets; (iii) Its global instance-level representations may be suboptimal for tasks like semantic segmentation that require local cues, necessitating adaptations of multi-scale learning or spatial constraints; (iv) Sensitive to the momentum coefficient used in target network updates.

\subsubsection{Prior-Guided Local (PGL)} 

\begin{figure}[!tp]
    \centering
    \begin{subfigure}[!tp]{\columnwidth}
        \centering
        \includegraphics[trim={21.2cm, 1cm, 9.2cm, 0.7cm}, clip, width=1\columnwidth]{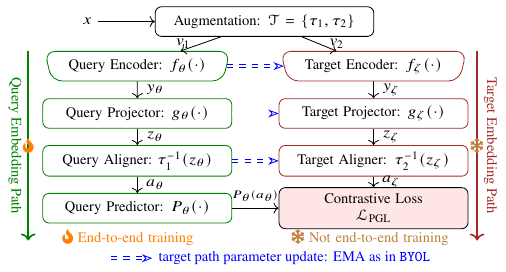}
        \caption{An illustration of the PGL pipeline inspired by Xie (2020) \cite{xie2020pgl}.}
        \label{fig:pgl-pipeline}
    \end{subfigure}\\
    \hfill
    \begin{subfigure}[!tp]{\columnwidth}
        \centering
        \includegraphics[trim={1cm, 0.1cm, 0.1cm, 0cm}, clip, width=0.95\columnwidth]{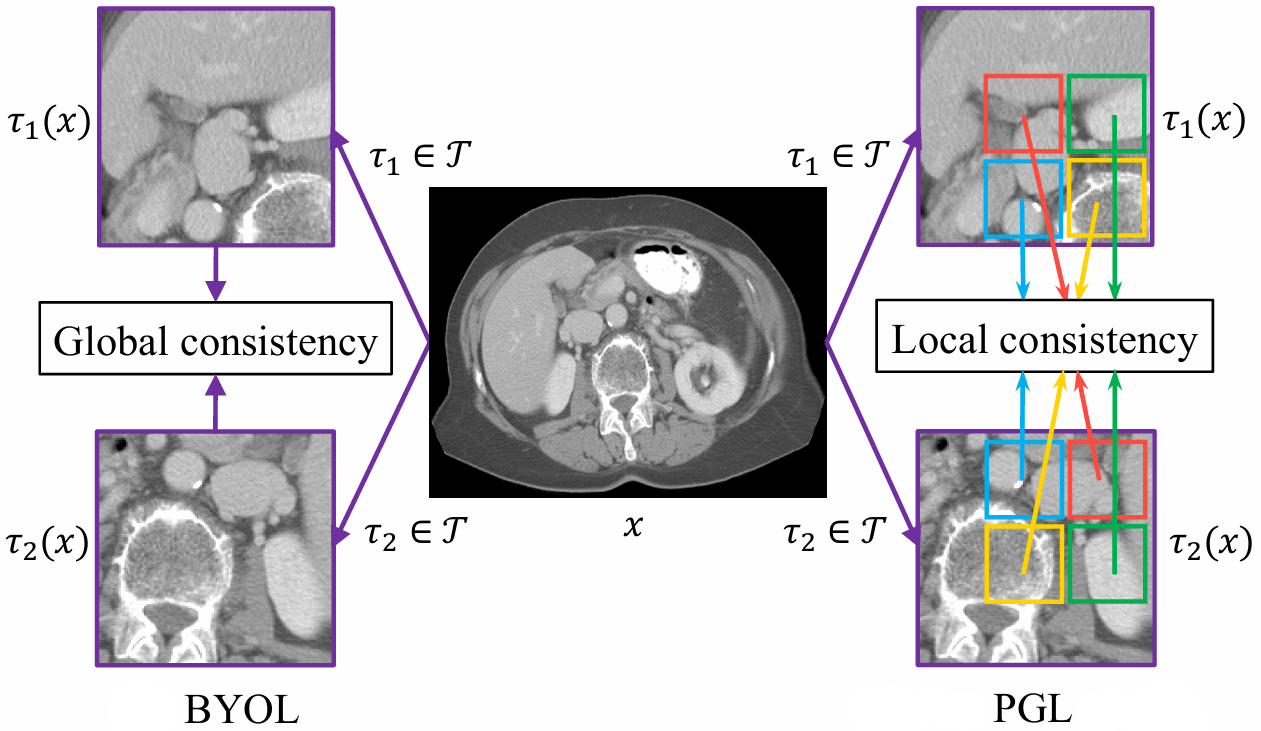}
        \caption{Concept of BYOL and PGL. This fig. is adopted from \cite{xie2020pgl}.}
        \label{fig:byolvspgl}
    \end{subfigure}
    \caption{The concept of the PGL. 
    }\label{fig:pgl} \vspace{-0.2cm}
\end{figure}

SimCLR and BYOL primarily focus on global feature consistency, often overlooking local feature alignment. However, learning local regional representations enhances the structural understanding, making models more effective for segmentation tasks, as illustrated in Fig.~\ref{fig:pgl}. To address this, Xie~\textit{et~al.}~\cite{xie2020pgl} introduced an additional \textit{aligner} layer after the projection head in both query and target paths, modifying the BYOL pipeline (cf.~Fig.~\ref{fig:pgl-pipeline}).
The aligner exploits augmentation information (e.g., flipping, cropping, scaling) to align features extracted from different views of the same region (cf.~Fig.~\ref{fig:byolvspgl}).
The augmentation module generates two views of an image: 
$v_1 = \tau_1(x)$ and $v_2 = \tau_2(x)$. The query and target paths produce feature representations: $z_\theta = g_\theta(f_\theta(v_1)) \text{, and } z_\zeta = g_\zeta(f_\zeta(v_2)),$
where $\theta$ and $\zeta$ are the parameters of the query and target networks, respectively. 

The alignment modules reverse the applied transformations:
$a_\theta = \tau_1^{-1}(z_\theta) \text{, and } a_\zeta = \tau_2^{-1}(z_\zeta),$
utilizing prior knowledge of $\tau_1$ and $\tau_2$. Ideally, these features represent the same local region of the input image. The predictor head $P_\theta(\cdot)$ in the query path then predicts the target aligner’s output. The model is optimized by minimizing the local consistency loss:
\begin{equation}
    \mathcal{L}_{\text{local}}(v_1, v_2) = \frac{\sum \|\mathcal{N}(P_\theta(a_\theta)) - \mathcal{N}(a_\zeta)\|_2^2}{K \times H \times W \times D},
    \label{eq:pgl-loss}
\end{equation}
where $\mathcal{N}$ represents $\ell_2$ normalization along the channel axis, and $K, H, W, D$ denote the batch size, height, width, and depth of the aligner features, respectively. Similarly, when $v_2$ is fed into the query network and $v_1$ into the target network, the loss is defined as $\mathcal{L}_{\text{local}}(v_2, v_1)$. The total loss minimized is:
\begin{equation}
    \mathcal{L}_{\text{PGL}} = \mathcal{L}_{\text{local}}(v_1, v_2) + \mathcal{L}_{\text{local}}(v_2, v_1).
    \label{eq:total-loss-pgl}
\end{equation}

Similar to BYOL, only the query path undergoes end-to-end training, while the target path is updated via EMA. 
PGL and its variants have demonstrated competitive performance in segmentation tasks across natural images, remote sensing, and 3D medical imaging~\cite{ye2024cads, li2023semantic, zheng2022msvrl, tian2020prior, xie2020pgl, heller2021state, xu2016multi}.

Despite its advantages, PGL has the following \textbf{limitations}:  
(i) The addition of aligner modules increases computational overhead, particularly for high-resolution images.  
(ii) It relies on augmentation priors and transformation reversals to align local features, which may not always be accurately defined in real-world applications.  
(iii) In datasets with fine-grained details, such as medical or remote sensing images, misalignment between augmented views can degrade feature consistency.  
(iv) Its performance is sensitive to hyperparameters, including patch sizes, augmentation strategies, and the trade-off between local and global consistency in the loss function. 


\subsubsection{Swapping Assignments between multiple Views (SwAV)}\label{sec-swav}

\begin{figure}[!tp]
\centering  
    \includegraphics[trim={30.45cm, 1.2cm, 0.05cm, 0.7cm}, clip, width=1.0\columnwidth]{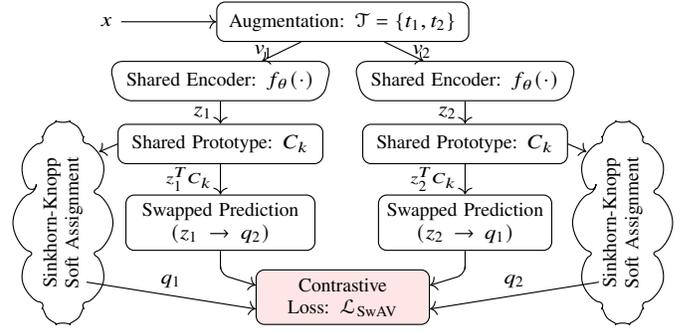}
    \caption{An illustration of the SwAV pipeline.}
    \label{fig:swav} \vspace{-0.2cm}
\end{figure}

Unlike MoCo, BYOL, or PGL, SwAV does not maintain separate query and target networks. Instead, it uses a shared network, akin to SimCLR, as shown in Fig.~\ref{fig:swav}. Unlike SimCLR, SwAV introduces a prototype layer in place of the projection layer and minimizes a unique loss function designed to dynamically balance cluster assignments~\cite{caron2020unsupervised}. 
It learns the input representations by comparing cluster assignments across different views rather than directly contrasting their features. Given an input $x$, random transformations $t_1$ and $t_2$ (e.g., cropping, aspect ratio changes) produce two augmented views, $v_1$ and $v_2$. Both views pass through a shared encoder $f_\theta(\cdot)$, yielding feature embeddings $z_1$ and $z_2$. These embeddings are then projected onto $K$ trainable prototype vectors, $C_k$, forming dot products $z_1^T C_k$ and $z_2^T C_k$. The resulting projections (prototype representations) are mapped to soft cluster assignments $\mathbf{q}_1$ and $\mathbf{q}_2$ via the Sinkhorn-Knopp algorithm~\cite{cuturi2013sinkhorn}.

SwAV’s core mechanism is swapped prediction: $\mathbf{z}_1^T C_k$ predicts $\mathbf{q}_2$, and $\mathbf{z}_2^T C_k$ predicts $\mathbf{q}_1$. Since both views originate from the same image, their features should encode similar cues, allowing cross-view prediction of cluster assignments. The training minimizes the discrepancy between predicted and actual assignments using a cross-entropy (CE) loss:
\begin{equation}
    l_{\text{CE}}{(\mathbf{z}_1, \mathbf{q}_2)} = - \sum_{k=1}^{K} \mathbf{q}_2^{(k)} \log \frac{\exp(\frac{1}{\tau}\mathbf{z}_1^T C_k)}{\sum_{k'=1}^{K} \exp(\frac{1}{\tau}\mathbf{z}_1^T C_{k'})},
    \label{eq:swav-loss}
\end{equation}
where $\tau$ is a temperature parameter controlling the sharpness of the softmax distribution. A similar loss $l_{\text{CE}}{(\mathbf{z}_2, \mathbf{q}_1)}$ is defined for the second view. Thus, the total loss function is:
\begin{equation}
    \mathcal{L}_{\text{SwAV}} = l_{\text{CE}}{(\mathbf{z}_1, \mathbf{q}_2)} + l_{\text{CE}}{(\mathbf{z}_2, \mathbf{q}_1)}.
    \label{eq:total-loss-swav}
\end{equation}
Minimizing $\mathcal{L}_{\text{SwAV}}$ jointly updates the encoder parameters, $\theta$, and the prototype matrix, $C$, allowing prototypes to adapt dynamically during training~\cite{caron2020unsupervised}.

However, one must be aware of SwAV's \textbf{limitations} before applying it to a problem domain: (i)  While effective on large-scale datasets (e.g., ImageNet), performance degrades on smaller datasets; (ii) Success depends on generating diverse and meaningful augmented views for robust cluster assignments; (iii) Clustering-based pseudo-labels struggle to capture fine-grained class distinctions in highly diverse datasets; (iv) Its global-level representation learning may not provide optimal features for dense prediction tasks without domain-specific adaptation; (v) The multi-crop augmentation strategy increases computational overhead during training.


\subsubsection{Simple Siamese Representation Learning (SimSiam)} \label{SimSiam}

\begin{figure}[!tp]
\centering  
    \includegraphics[trim={32.6cm, 3.5cm, 0.1cm, 1.0cm}, clip, width=1\columnwidth]{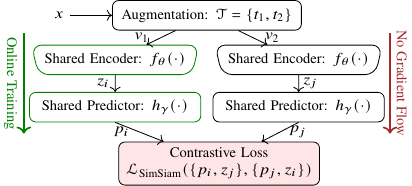}
    \caption{SimSiam's overview based on pseudocode of~\cite{chen2021exploring}.}
    \label{fig:SimSiam}\vspace{-0.2cm}
\end{figure}

Fig.~\ref{fig:SimSiam} illustrates the workflow of SimSiam, a simplified variant of SimCLR~\cite{chen2020simple}, BYOL~\cite{grill2020bootstrap}, and SwAV~\cite{caron2020unsupervised}. 
It has several distinguishing features:
(i) Unlike MoCo, BYOL, and PGL, SimSiam removes the momentum encoder, relying solely on shared weights for both branches.
(ii) Like SimCLR and SwAV, it shares the encoder and predictor across both branches but introduces asymmetry via a stop-gradient applied to one branch.
(iii) Unlike SimCLR, SimSiam does not use $-$ve pairs, and unlike SwAV, it does not involve clustering or prototypes. Instead, it relies solely on $+$ve pairs and the stop-gradient operation to prevent collapse.

Given an input $x$, two augmented views $v_1$ and $v_2$ are generated. A shared encoder $f_\theta(\cdot)$ processes these views, yielding feature representations $z_i$ and $z_j$. A shared predictor then maps $z_i$ to a prediction $p_i = h(z_i)$. Training minimizes the negative cosine similarity between $p_i$ and $z_j$:
\begin{equation}
S_{cos}(p_i, z_j) = - \frac{p_i}{\|p_i\|_2} \cdot \frac{z_j}{\|z_j\|_2},
\end{equation}
where $\|\cdot\|_2$ denotes the $\ell_2$-norm. The total symmetric loss is:
\begin{equation}
\mathcal{L} = \frac{1}{2} S_{cos}(p_i, z_j) + \frac{1}{2} S_{cos}(p_j, z_i),
\end{equation}
averaged over all samples. The online branch is trained via backpropagation, while the other branch remains fixed (i.e., no gradient updates or momentum-based updates).

SimSiam has been used to pre-train models, enabling the extraction of transferable features for various downstream tasks. In the original study~\cite{chen2021exploring}, SimSiam was evaluated on object detection and instance segmentation benchmarks, including Pascal VOC~\cite{everingham2010pascal} and MS COCO~\cite{lin2014microsoft}. Results demonstrated competitive performance against contemporary SSL methods, highlighting SimSiam’s effectiveness.
Its advantages—particularly the elimination of $-$ve pairs and its ability to train with small batch sizes—make it well-suited for semantic segmentation~\cite{zou2022spot, wang2024multi, gandhi2024self, sakaguchi2024object}. For example, Wang \textit{et al.}~\cite{wang2024multi} applied SimSiam to histopathology image segmentation, reporting superior performance over other SSL approaches. Similarly, Gandhi \textit{et al.}~\cite{gandhi2024self} used SimSiam with a ResUNet++~\cite{jha2019resunet++} backbone for land cover segmentation in satellite imagery. Their findings show that SimSiam outperforms SimCLR when using only $50\%$ of labeled data. Additionally, Sakaguchi~\textit{et al.}~\cite{sakaguchi2024object} integrated SimSiam into an object instance retrieval framework for assistive robotics, exploiting 3D semantic map views to enhance performance.

\textbf{Limitations} of SimSiam are: (i) The absence of negative samples makes performance highly sensitive to architecture and data augmentation choices, with suboptimal configurations leading to degraded results.
(ii) The stop-gradient mechanism, while preventing representation collapse, may require extensive fine-tuning for specific tasks, adding computational overhead. (iii) It has mainly been validated on modest backbone architectures; its scalability to large datasets and high-capacity models (e.g., transformers) remains underexplored.


\subsection{The Emerging Ideas}\label{sec-current-trend}

Numerous studies have demonstrated the effectiveness of self-supervised learning (SSL) in image segmentation tasks, achieving performance comparable to fully supervised methods while significantly reducing the need for labeled data.  
However, most SSL approaches are designed to learn global representations, which may be suboptimal for dense prediction tasks, like semantic segmentation~\cite{liu2025unsupervised, zheng2022msvrl}.
To address this limitation, recent advancements focus on dense SSL approaches, either at the pixel-level~\cite{dhamale2025dual, wang2021dense, xie2021propagate} or region-level~\cite{xiao2021region, yang2013saliency}. 
For instance, {DenseCL}~\cite{wang2021dense} extends MoCo v2~\cite{chen2020improved} by introducing dense pairwise contrastive learning at the pixel-level. 
{DenseSiam}~\cite{zhang2022dense} employs a dense Siamese network that exploits both pixel and region consistency for enhanced dense feature learning.
Chaitanya~\textit{et al.}~\cite{chaitanya2020contrastive} come up with a local contrastive loss that enforces similarity between transformed local regions while ensuring dissimilarity with other areas of the same image.
The local contrastive loss for a given similar image pair, $\{\tilde{x}, \hat{x}\}$ of size $W_1 \times W_2 \times C$, is defined as:
\begin{equation}
    l(\tilde{x}, \hat{x}, u, v) = - \log \frac{\text{exp}[\text{sim}(\tilde{f}(u,v), \hat{f}(u,v)) / \tau]}
    {\text{exp}[\text{sim}(\tilde{f}(u,v), \hat{f}(u,v)) / \tau] + \Gamma},
\end{equation}
where $\text{sim}(\cdot,\cdot)$ is the cosine similarity, $\tau$ is a temperature parameter---a smaller $\tau$ emphasizes the difference in similarity, which helps the loss focus more on the most similar pairs and the hardest dissimilar pairs. $(u, v)$ indexes local regions in the feature maps, $f(u,v) \in \mathbb{R}^{K \times K \times C}$, forming the similar pair set $\Omega^+$, and $\Gamma$ is defined as:
\[ \Gamma = \sum_{(u',v') \in \Omega^-} \text{exp}[\text{sim}(\tilde{f}(u,v), \hat{f}(u',v')) / \tau],\]
where $\Omega^-$ denotes the dissimilar set consisting of all other local regions in both feature maps, $\tilde{f}$ and $\hat{f}$.
Thus, the total local contrastive loss, when each feature map is divided into $D$ local regions, each of size $K \times K \times C$, with $K < \min(W_1, W_2)$, for a set of images from $\mathbb{X}$, can be defined as:
\begin{equation}
    L_l = \frac{1}{|\mathbb{X}|} \sum_{x \in \mathbb{X}} \frac{1}{D} \sum_{(u,v) \in \Omega^+} \Big[l(\tilde{x}, \hat{x}, u, v) \Big].
\end{equation}

Experimental studies demonstrate that incorporating local contrastive loss improves MRI image segmentation over traditional approaches~\cite{chaitanya2020contrastive, dhamale2025dual}. Likewise, Xie~\textit{et~al.}~\cite{xie2020pgl} explore region-wise local consistency in the latent space of SSL for medical image semantic segmentation. Islam~\textit{et~al.}~\cite{Islam_2023_WACV} introduce a pixel-level contrastive loss into BYOL, enhancing dense feature representations for downstream tasks of object detection and semantic segmentation. 
Another work in \cite{ye2024cads} proposes a novel SSL framework for medical image segmentation, termed cross-modal alignment and deep self-distillation (CADS). This method integrates concepts from CPC and BYOL, formulating a hybrid SSL approach. The cross-modal alignment component utilizes a differentiable projection module to generate digitally reconstructed radiographs (DRRs) from 3D CT volumes, serving as X-ray proxies. This alignment enables effective feature learning across 2D and 3D modalities. Meanwhile, the deep self-distillation component employs a teacher-student framework, refining hierarchical feature representations similar to BYOL. 
By aligning features across modalities, CADS enhances modality-invariant feature learning, which is crucial for volumetric medical image segmentation that requires spatial and contextual understanding in 3D data. While CADS achieves promising results on CT scan segmentation, its performance remains sensitive to input image/patch size and lacks generalizability across different imaging modalities and tasks.

\subsection{Summary}

Table~\ref{tab:pretext-literature-summary} and Table~\ref{tab:comparison-of-key-SSL-methods} summarize the details of the three categories of SSL--predictive, generative, and contrastive learning--pretext tasks, offering a concise breakdown of their characteristics and applications. 
Furthermore, Table~\ref{tab:comparison-of-key-SSL-methods} provides a comparative overview of the contrastive SSL methods discussed in this study, evaluated in eight key properties.

\begin{table*}[!tp] 
\centering
\setlength{\tabcolsep}{6pt}
\renewcommand{\arraystretch}{1.1} 
\caption{Comparison of popular predictive and generative self-supervised learning approaches.} \label{tab:pretext-literature-summary}
\begin{tabular}{m{0.12cm} m{2.5cm}m{7.0cm} m{6.5cm} } 
\hline 
\textbf{} & \textbf{Pretext Task} & \textbf{Description} & \textbf{Key Characteristics}  \\ 
\hline \hline
\parbox[t]{1mm}{\centering\multirow{9}{*}{\rotatebox[origin=c]{90}{\textcolor{Brown}{Predictive SSL}}}} 
& Slice Order & Divide an image into random slices for reordering & Simple to implement but sensitive to slice order, leading to artifacts \\ \cline{2-4}
& Jigsaw Puzzle & Divides an image into patches, rearranges them randomly, and trains the model to predict the original positions, enhancing spatial relationship learning & Improves spatial understanding but requires high computation and large datasets \\ \cline{2-4}
& Rubik’s Cube & Simulates a 3D Rubik’s Cube by rotating and permuting image patches, training the model to predict both transformations & Trains the model on spatial and orientation-aware features, but computational complexity increases significantly due to the 3D nature of the task \\ \cline{2-4}
& Rotation Prediction & Rotates an image along horizontal axes and trains the model to predict the rotation, enhancing orientation-invariant learning & Simple and effective for orientation learning, but limited to discrete rotations \\ \hline 

\parbox[t]{1mm}{\centering\multirow{9}{*}{\rotatebox[origin=c]{90}{\textcolor{Green}{Generative SSL}}}} 
& Denoising & Learns to reconstruct clean images from noisy inputs, encouraging the model to extract robust features & Utilizes large unlabeled datasets; effective for noise removal but sensitive to noise types and levels \\ \cline{2-4}
& Colorization & Predicts realistic colors for grayscale images by learning contextual and semantic features & enhances understanding of object semantics but may require post-processing for accurate results \\ \cline{2-4}
& Inpainting & Predicts missing or occluded regions in an image by leveraging surrounding contextual information & Improves spatial and contextual feature learning; effective for object removal and photo restoration but may struggle with complex or large missing areas \\ \cline{2-4}
& Context Restoration & Trains the model to restore or predict spatial relationships and contextual consistency within an image & Enhances the model's ability to capture object relationships and spatial structure; may require complex training procedures \\ \hline 
\parbox[t]{1mm}{\centering\multirow{7}{*}{\rotatebox[origin=c]{90}{\textcolor{Purple}{Contrastive SSL}}}} 
& CPC & Maximizes mutual information across views & State-of-the-art results; limited spatial focus  \\ \cline{2-4}
& SimCLR & Maximizes intra-image similarity across views & Versatile but computationally expensive  \\ \cline{2-4}  
& MoCo & Uses memory bank for feature updates & Efficient memory use; high data requirements \\ \cline{2-4}
& PGL & Learns graph-based patch relationships & Captures spatial relations; costly for large datasets  \\ \cline{2-4}
& BYOL & Uses consistency loss across augmentations & Memory-efficient; sensitive to hyperparameters  \\ \cline{2-4}
& SwAV & Multi-crop augmentations for representation learning & Good for transfer learning; computationally heavy  \\ \cline{2-4}
& SimSiam & Two-branch network with stop-gradient & Simple, scalable; less effective for low-diversity datasets \\ 
\hline \hline

\end{tabular}
\end{table*}

\begin{table*}[!tp]
\setlength{\tabcolsep}{3pt} 
\centering
\caption{Comparison of popular contrastive learning approaches based on key features.} \label{tab:comparison-of-key-SSL-methods}
\renewcommand{\arraystretch}{1.5} 
\begin{tabular}{>{\raggedright\arraybackslash}m{1.5cm}%
                >{\raggedright\arraybackslash}m{2.0cm}%
                >{\raggedright\arraybackslash}m{2.5cm}%
                >{\raggedright\arraybackslash}m{2.5cm}%
                >{\raggedright\arraybackslash}m{2.9cm}%
                >{\raggedright\arraybackslash}m{2.5cm}%
                >{\raggedright\arraybackslash}m{2.5cm}}
\hline
\textbf{Feature} & \centering\texttt{\textbf{SimCLR}} & \centering\texttt{\textbf{MoCo}} & \centering\texttt{\textbf{BYOL}} & \centering\texttt{\textbf{PGL}} & \centering\texttt{\textbf{SwAV}} & \quad\quad\texttt{\textbf{SimSiam}}\\ \hline \hline

\textbf{Methodology} 
& Contrastive learning with large batches 
& Momentum contrast mechanism 
& Negative-sample-free, self-distillation, momentum update 
& Prior-guided local alignment for segmentation tasks 
& Online clustering with Sinkhorn-Knopp algorithm 
& Negative-sample-free, no momentum update, stop-gradient \\ \hline

\textbf{Architecture} 
& Encoder + Projection 
& Encoder + Memory Queue 
& Student + Teacher 
& Aligner modules with query/target paths 
& Encoder + Prototype layer 
& Encoder + Projection head \\ \hline

\textbf{Learning Objective} 
& Contrastive loss 
& Contrastive loss 
& Self-distillation 
& Local and global consistency loss 
& Swapped prediction with clustering 
& Self-distillation \\ \hline

\textbf{Negative Samples} 
& Required (minibatch) 
& Required (memory bank) 
& Not required 
& Not required 
& Not required (clustering instead) 
& Not required \\ \hline

\textbf{Dependence on Batch} 
& High 
& Low (memory bank) 
& None 
& Medium 
& Low 
& Low \\ \hline

\textbf{Simplicity} 
& Moderate 
& Complex 
& Simple 
& Moderate 
& Moderate 
& Simple \\ \hline

\textbf{Strengths} 
& Simplicity, scalable 
& Memory-efficient, suitable for smaller batches 
& No negative samples, avoids batch size constraints 
& Captures local structural information, ideal for segmentation 
& Balanced clustering, robust to batch sizes 
& Easy to implement, small batch sizes, no negative samples \\ \hline

\textbf{Limitations} 
& Requires large batch sizes 
& Requires a memory queue 
& Sensitive to augmentation strategy 
& Computational overhead due to aligners; depends on priors 
& Sensitive to clustering hyperparameters 
& Sensitive to architecture and augmentation \\ \hline\hline

\end{tabular}
\end{table*}

\section{Commonly Used Datasets} \label{dataset}

\begin{table*}[!ht]
\centering
\setlength{\tabcolsep}{3.0pt} 
\renewcommand{\arraystretch}{1.2} 
    \caption{Widely used publicly available datasets for semantic segmentation research and development.}\label{tab:dataset} 
    \begin{tabular}{p{2.4cm}|p{11.3cm}| p{1.5cm}|p{1.1cm}| >{\centering\arraybackslash}p{0.7cm}}  \hline

\centering \vspace{0cm}\textbf{Dataset} & \centering \vspace{0cm}\textbf{Description} & \centering \textbf{Image \newline Resolution} & \centering \textbf{\# of Samples} & \textbf{\# of Classes} \\  \hline\hline

KiTS 2023 \cite{heller2023kits21} & Kidney CT scans with a ternary mask identifying semantic segmentation of kidneys, renal tumors, and renal cysts. 489 cases are allocated to the training and 110 cases are for testing sets. &  Variable & 599 & 3  \\  

BraTS 2024 \cite{labella2024brain} & Brain MRIs with detailed voxel-level annotations for enhancing tumor, non-enhancing tumor core, and surrounding FLAIR hyper-intensity, consisting of 500, 70, and 180 samples for training, validation, and testing, respectively. & Variable & 750 & 3 \\  

ISIC 2018 \cite{codella2019skin} &	DICOM skin imaging for skin lesion segmentation and classification. Image-level labels and pixel-level segmentation masks are provided for lesion segmentation (3,694) and 5-way lesion attribute detection (3,694), and 7-way lesion disease classification (11,720). & $1024\times1024$ & 18,108 & 2/5/7  \\ 

Cityscapes \cite{cordts2016cityscapes} & Urban street scenes with detailed dense pixel-level annotations for various objects. & $2048\times1024$ & 5,000 & 30 \\  

CamVid \cite{brostow2009semantic} & Urban scene understanding dataset with pixel-wise semantic segmentation. & $960\times720$& 701 & 32 \\ 


SYNTHIA \cite{ros2016synthia} & Photo-realistic synthetic urban scenes of random snapshots and video sequences. Includes different seasons, weather, and illumination conditions from multiple viewpoints with pixel-level annotations. & $1280\times960$ & $213,400+$ & 13 \\

ADE20K \cite{xie2021selfsupervised}	& A comprehensive dataset with 150 object semantic segmentation masks, with 20,000 images for training, 2,000 images for validation, and 2,000 images for testing. & Variable & $20,000+$ & 150 \\  

PASCAL VOC	\cite{everingham2015pascal} & Popular dataset for object recognition and segmentation with pixel-level annotations. & Variable & 11,530 & 20 \\

MS COCO \cite{lin2014microsoft} & Contains detailed pixel-level masks for 80 diverse objects in various backgrounds. & Variable & 330,000 & 80 \\

\hline \hline

\end{tabular}
\end{table*}

The advancement of image segmentation has been driven by benchmark datasets released by various groups and institutions that facilitate model development, evaluation, and standardized analysis.
This section summarizes commonly used datasets and their properties. 
Table~\ref{tab:dataset} provides an overview of some publicly available benchmark datasets for segmentation. 

\subsection{Medical Imaging Datasets}

\textbf{KiTS 2023~\cite{heller2023kits21}:} 
It is the 3rd iteration of the Kidney Tumor Segmentation (KiTS) Challenge organized by the University of Minnesota, the German Cancer Research Center (DKFZ), and the Cleveland Clinic's Urologic Cancer Program. 
It provides high-quality segmentation annotations of kidneys, kidney tumors, and kidney cysts composed of 599 cases (training set - 489 and test set - 110). 
The current SOTA model achieves a Dice score of $83.5\%$ according to \href{https://kits-challenge.org/kits23/#kits23-official-results}{KiTS leaderboard}.

\textbf{BraTS 2024~\cite{labella2024brain}:} 
It is the 13th iteration of the long-standing Brain Tumor Segmentation (BraTS) Challenge, which has been held annually since 2012. Over the years, BraTS has expanded its scope to address various clinical concerns and tumor types as part of its ongoing efforts to advance automated brain tumor segmentation.
It provides a labeled dataset consisting of 500, 70, and 180 samples for training, validation, and testing, respectively. However, researchers can utilize previous versions that contain additional training samples to enhance their model performance.
The current SOTA model achieves a Dice score of $89.4\%$ according to the \href{https://www.synapse.org/Synapse:syn53708249/wiki/628815}{BraTS leaderboard}.

\textbf{ISIC 2018~\cite{codella2019skin}:} 
The 2018 International Skin Imaging Collaboration (ISIC) Challenge is the 3rd iteration, held at MICCAI, aimed to advance AI-driven dermatological analysis. It has 3 tasks:
(i) \textit{Binary lesion segmentation}: Training set contains 2,594 dermoscopic images with segmentation masks. The validation and test sets include 100 and 1,000 images, respectively, without ground truth masks.
(ii) \textit{5-way lesion detection}: It involves detecting 5 dermoscopic attributes. The training set has 2,594 images w/t 12,970 segmentation masks (5 per image). The validation and test sets have 100 and 1,000 images, respectively, without ground truths.
(iii) \textit{7-way lesion classification}: Training set has 10,015 images w/t labels for 7 disease types. The validation and test sets have 193 and 1,512 images, respectively, without ground truths.
The \href{https://challenge.isic-archive.com/leaderboards/2018/}{ISIC 2018 leaderboard} shows the model that performed the best in the segmentation task and achieved a Jaccard index of $80.2\%$. In later iterations, the focus shifted toward classification tasks.

\subsection{Urban Scene Understanding Datasets}

\textbf{CamVid \cite{brostow2009semantic}:} 
The Cambridge-driving Labeled Video Database (CamVid) is a pioneering dataset for semantic segmentation in driving scenarios. It consists of 5 video sequences captured using a $960\times720$ resolution camera mounted on a vehicle's dashboard. From these sequences, 701 frames were manually annotated, with each pixel assigned to one of the 32 semantic classes.
A common split includes 367 training, 101 validation, and 233 test images. 
According to \href{https://paperswithcode.com/sota/semantic-segmentation-on-camvid}{recent evaluations}, the SOTA model SERNet-Former has achieved a mean intersection over union (mIoU) of $84.6\%$.


\textbf{Cityscapes \cite{cordts2016cityscapes}:}  
It is a large-scale dataset for semantic and instance segmentation in urban street scenes, widely used in autonomous driving research. It helps in understanding the driving environment, pedestrian detection, and traffic analysis, supporting self-driving cars and intelligent transportation systems. The dataset includes high-resolution images from 50 cities, covering diverse urban environments with 30 semantic classes. It provides 5,000 finely annotated images and 20,000 coarsely annotated images, enabling both detailed analysis and large-scale training. Coarse annotations can be used for pre-training, followed by fine-tuning on finely annotated images to improve model performance.
The current state-of-the-art model on the Cityscapes benchmark achieves a mIoU of $86.7\%$ according to \href{https://www.cityscapes-dataset.com/benchmarks/#scene-labeling-task}{Cityscapes leaderboard}.
Its latest version provides 3D bounding box annotations for all vehicles in the original Cityscapes samples and a benchmark for 3D detection.

\textbf{SYNTHIA 2016 \cite{ros2016synthia}:} The SYNTHetic collection of Imagery and Annotations (SYNTHIA) is a dataset designed to aid semantic segmentation and scene understanding in driving scenarios. It comprises over 200,000 high-definition images from video streams and more than 20,000 HD images from independent snapshots, all rendered from a virtual city with precise pixel-level semantic annotations for 13 classes. 
This dataset covers various conditions lighting, and weather (dynamic lights, shadows, multiple daytime modes, rain, etc.). 
The SOTA model, \href{https://paperswithcode.com/sota/semantic-segmentation-on-synthia-cvpr16}{SSMA}, achieves an mIoU of $91.3\%$ on the SYNTHIA dataset. The latest version, SYNTHIA-AL 2019, extends the number of classes to 15 and provides additional annotations, including instance segmentation, 2D and 3D bounding boxes, and depth information.

\subsection{General Purpose and Specialized Datasets}

\textbf{ADE20K \cite{xie2021selfsupervised}:} 
It is a comprehensive collection of over 27,000 images sourced from the SUN and Places databases, meticulously annotated to include more than 3,000 object categories. The dataset is divided into 25,574 images, and 2,000 images for training and validation, respectively. 
As of the \href{https://paperswithcode.com/sota/semantic-segmentation-on-ade20k}{latest evaluations}, the SOTA model on the ADE20K dataset is ONE-PEACE with an mIoU of $63\%$.

\textbf{PASCAL VOC \cite{everingham2015pascal}:} 
The PASCAL Visual Object Classes (VOC) dataset, introduced in 2005, has been a cornerstone in advancing object recognition and semantic segmentation research. The 2012 iteration, PASCAL VOC 2012, comprises 11,530 images annotated with 20 object categories, including vehicles, animals, and household items. Over the years, PASCAL VOC 2012 has served as a benchmark for various semantic segmentation models. Notably, SegNext achieved an mIoU of $90.60\%$ according to \href{http://host.robots.ox.ac.uk:8080/leaderboard/displaylb_main.php?challengeid=11&compid=6}{PASCAL VOC leaderboard}.

\textbf{MS COCO \cite{lin2014microsoft}:} 
The Microsoft Common Objects in Context (MS COCO) dataset is a comprehensive resource extensively utilized in computer vision research, particularly for semantic segmentation. It comprises over 330,000 images, with more than 200,000 labeled, encompassing 1.5M object instances across 80 object categories. Each image is annotated with pixel-level segmentation masks. 
In recent advancements, the \href{https://paperswithcode.com/sota/instance-segmentation-on-coco}{Co-DETR} model has achieved SOTA performance with an mIoU of $57.10\%$ on the COCO test benchmark for instance segmentation. This model leverages a transformer-based architecture to enhance object detection and segmentation capabilities, demonstrating significant improvements in accuracy.

\section{Challenges and Future Research Directions}\label{sec-future-challenges}

\subsection{Challenges}

Self-supervised image segmentation is an emerging research area in computer vision that seeks to address the limitations of traditional supervised segmentation methods, which rely on large amounts of labeled data.  
Although SSL has made significant progress, identifying an optimal pretext task tailored to specific downstream segmentation applications remains challenging. The \textbf{challenges} can be summed as: 

\noindent(i) \textit{Limited labeled data} - Since self-supervised methods do not rely on annotations, benchmarking and comparing different approaches is difficult due to the scarcity of labeled datasets. (ii) \textit{Defining meaningful objectives} - Designing self-supervised objectives that effectively capture structural patterns while remaining generalizable across datasets is challenging. 

\noindent (iii) \textit{Sensitivity to initialization and hyperparameters} - Training stability and performance are highly dependent on initialization and hyperparameter tuning, leading to potential instability. (iv) \textit{Capturing complex relationships} - Many self-supervised approaches primarily employ local and low-level features, limiting their ability to model high-level contextual relationships. (v) \textit{Interpretability and generalization} - Self-supervised models often lack interpretability and may struggle to generalize to unseen data, restricting their practical applicability.

\subsection{Future Research Directions}

In light of the above challenges, self-supervised image segmentation remains an active research area, offering promising opportunities for improvement and innovation.
For instance, \cite{deng2021does} proposed a network, integrating SSL with supervised learning, with a shared backbone for both ways of learning: one branch performing 4-way rotation prediction as in SSL and another conducting $n$-way classification as in supervised learning, where the backbone is optimized using a combined loss from both tasks. More similar innovations and continued research are necessary to advance self-supervised segmentation. 
Here are some potential \textbf{future research directions}:

\noindent (i) \textit{Incorporating Semantic Information} - Most self-supervised segmentation methods focus on predicting spatial relationships within an image. Incorporating semantic information (e.g., object labels) into the SSL process could enhance accuracy and interpretability.

\noindent (ii) \textit{Domain Adaptation} - Self-supervised segmentation models are typically trained on large datasets and then applied to unseen data. However, performance may degrade when applied to datasets with domain shifts. Developing domain adaptation techniques could improve model generalization.

\noindent (iii) \textit{ Few-Shot/ Zero-Shot Learning} - Extending the models to operate in few-shot and zero-shot settings would enable them to segment objects with minimal or no labeled examples, broadening their practical applicability.

\noindent (iv) \textit{Weakly-Supervision} - Combining SSL with weak supervision, where a small amount of labeled data guides the learning process, could enhance segmentation results.

\noindent (v) \textit{Interactive Methods} - Developing self-supervised models capable of incorporating user feedback could improve segmentation tools, enhancing their practicality.

\noindent (vi) \textit{Real-Time Segmentation} - Many applications, viz. robotics and autonomous vehicles, require real-time segmentation. Optimizing SSL segmentation models would expand their applicability in time-sensitive environments.

\section{Conclusion}\label{conclusion}

This study investigates recent developments in self-supervised learning for image segmentation.  
While supervised learning remains the standard, its dependence on large annotated datasets has increased interest in self-supervised learning as an efficient alternative.  
By categorizing pretext tasks and examining dataset sources, this work comprehensively analyzes self-supervised learning's potential across various domains.

Key insights from this study include: (i) Scalability - SSL segmentation methods enable training on large datasets without manual annotation. 
(ii) Diverse Applications: SSL-driven segmentation extends beyond medical imaging to agriculture, robotics, and other fields.
(iii) Multitasking: Integrating multiple pretext tasks enhances feature learning and representation quality of the models for downstream tasks.

As SSL continues to evolve, future research should prioritize efficiency, scalability, and adaptability to complex segmentation challenges. Advancements in these areas will be crucial for maximizing the utility of unlabeled data in building robust models. Our ongoing efforts focus on refining SSL methodologies for domain-specific applications, reinforcing its role as a transformative force in machine learning.

\bibliographystyle{ieeetr} 
\bibliography{references}

\vspace{12pt}


\onecolumn

\section{\textbf{Appendix}}

This section provides additional details relevant to the article titled ``Self-Supervised Learning for Image Segmentation: A Comprehensive Survey.'' 

\subsection{Organization Mind map}

\begin{figure*}[!hbp]
    \centering
          \begin{tikzpicture}[node distance=0.2cm, auto]
      \tikzstyle{block} = [rectangle, draw, fill=blue!20, text width=2.5cm, text centered, rounded corners=4pt, minimum height=3em]
      \tikzstyle{line} = [draw, -latex']
    
      \node [block, minimum width=2.5cm, fill=green!5] (block1) {Introduction};
      \node [block, right=of block1, minimum width=2.5cm, fill=Violet!15] (block2) {Existing Survey};
      \node [block, right= 3.2cm of block2, fill=pink!25] (block4) {Main Text};
      \node [block, right=of block4, xshift=2.7cm, fill=blue!10] (block5) {Conclusion};
      \node [block, below=of block1, align=left, minimum width=2.6cm, text width=2.6cm, minimum height = 3.1cm, yshift=-0.2cm, fill=green!5, draw=green!50] (block6) { $\bullet$ Problem statement, and motivation. \newline $\bullet$ Traditional vs. modern machine learning. \newline $\bullet$ Contributions.};
      \node [block, below=of block2, align=left, minimum width=2.6cm, text width=2.6cm, minimum height = 3.1cm, yshift=-0.2cm, fill=brown!5, draw=brown!50] (block7) {$\bullet$ Tracking previous studies on SSL for image segmentation. $\bullet$ Comparison of the previous studies.};
      \node [block, below=of block2, xshift=2.9cm, yshift=-0.2cm, fill=pink!15, draw=pink!50] (block8) {Preliminary};
      \node [block, below=of block8, fill=pink!15, draw=pink!50, minimum height = 1.6cm, minimum width=2.45cm, text width=2.45cm] (block9) {Basic concepts of SSL and\\ image segmentation.};
      \node [block, below=of block4, xshift=0cm, yshift=-0.2cm, minimum width=2.9cm, text width=2.9cm, fill=pink!10, draw=pink!50] (block10) {Pretext tasks for image segmentation};
      \node [block, below=of block10, minimum height = 1.6cm, minimum width=2.9cm, text width=2.9cm, fill=pink!10, draw=pink!50] (block11) {Categorization of pretext tasks for better understanding.};
      \node [block, right= 0.1cm of block10, fill=pink!5, draw=pink!50, minimum width=2.2cm, text width=2.2cm] (block12) {Benchmark Datasets};
      \node [block, below=of block12, fill=pink!5, draw=pink!50, minimum width=2.2cm, text width=2.2cm, minimum height=1.8cm] (block13) {Detailed description on the benchmark datasets.};
      \node [block, below=of block5, yshift=-0.2cm, minimum height=3.1cm, align= left, fill=blue!10, draw=blue!50] (block14) {$\bullet$ Challenges in SSL-based image segmentation.\newline $\bullet$ Final thoughts and future direction of SSL.};
    
      \draw [line] (block1) -- (block2);
      \draw [line] (block2) -- (block4);
      \draw [line] (block4) -- (block5);
      \draw [dashed] (block1) -- (block6);
      \draw [dashed] (block2.south) .. controls +(0,-0.5) and +(-0.1,0.5) .. (block7.north);
      \draw [line] (block4) -- (block8);
      \draw [dashed] (block8) -- (block9);
      \draw [line] (block4) -- (block10);
      \draw [dashed] (block10) -- (block11);
      \draw [line] (block4) -- (block12);
      \draw [dashed] (block12) -- (block13);
      \draw [dashed] (block5) -- (block14);
     
    \end{tikzpicture}   
        \caption{The mind map of the survey. }
        \label{fig:concept-diagram} \vspace{-1cm}
    \end{figure*}
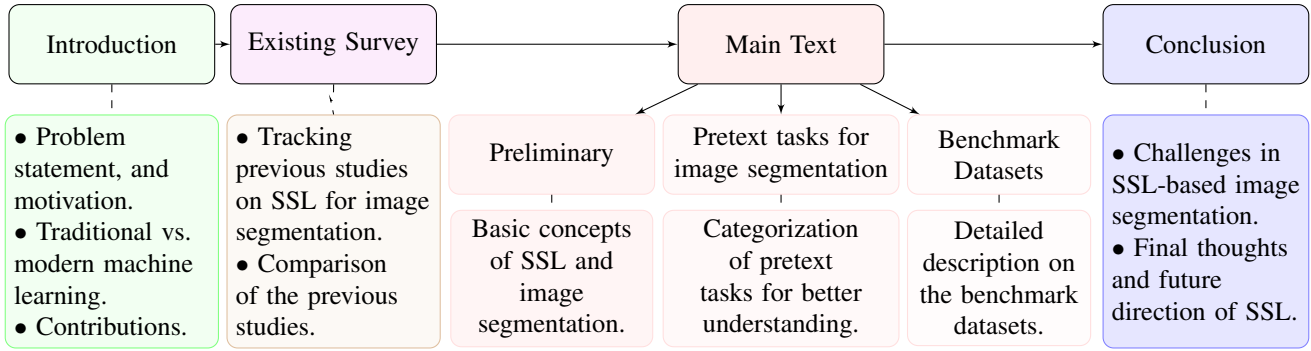

Fig.~\ref{fig:concept-diagram} presents a mind map of this study.

\subsection{Research Methodology}
    
\begin{figure}[htp]
        \centering
    \begin{tikzpicture}[node distance=1.5cm]
    
    \node (start) [process-start, xshift=0cm, text width=8.5cm, minimum width=8.5cm, inner sep=0.1cm] {\textbf{Article Searching on Academic Databases}\\ \raggedright\textit{Search terms}: {image segmentation,  pretext tasks, self-supervised image segmentation, self-supervised learning, self-supervised learning survey, self-supervised learning review} \\\textit{Collected publications}: PubMed - 25, Science Direct - 50,  IEEE Xplore - 45, ACM Digital Library - 75, Google Scholar - 55\\ };
    \node (store) [storage, below of=start,  yshift=-0.6cm,  xshift=-2.2cm, text width=3.5cm, minimum width=3.5cm, fill=blue!10] {\textbf{Storing All Uncurated Articles}\\ Number of articles - 250};
    \node (filter1) [process-optional, right of=store, yshift=0cm, xshift=2.8cm, text width=4.2cm, minimum width=4.2cm, fill=green!5] {\textbf{Filtering Redundancies} \\ Removed duplicates = 84};
    \node (curate) [startstop, below of=store, yshift=0cm, xshift=0cm, text width=3.5cm, minimum width=3.5cm, fill=blue!5] {\textbf{Article Curation} \\ Organized articles according to title and abstract - 166};
    \node (filter2) [process-optional, right of=curate, yshift=0cm, xshift=2.8cm, text width=4.2cm, minimum width=4.2cm, fill=green!5] {\textbf{Filtering Irrelevancies}\\ Excluding the articles that are not suitable for survey structure or that do not include full text.};
    \node (review) [startstop, below of=start, yshift=-3.6cm, xshift=0cm, text width=8.5cm, minimum width=8.5cm, fill=red!5] {\textbf{Reviewing Articles}: 166 curated set of articles for survey};
    \draw [arrow] (start.south) .. controls +(down:2mm) and +(up:5mm) .. (store.north);
    \draw [dashed] (store) -- (filter1);
    \draw [arrow] (store) -- (curate);
    \draw [dashed] (curate.east) .. controls +(right:0mm) and +(up:0mm) .. (filter2.west);
    \draw [arrow] (curate.south) .. controls +(down:2mm) and +(up:5mm) .. (review.north);
    
    \end{tikzpicture}
     \caption{The article searching and selection procedure.}
    \label{fig:flowchart}
    \end{figure}
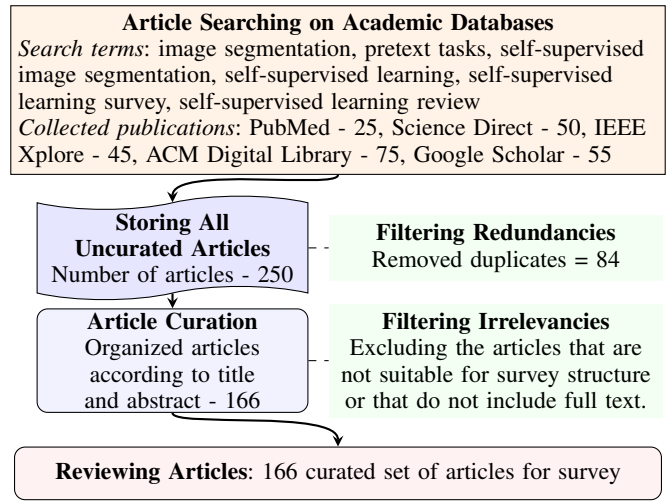

    Fig.~\ref{fig:flowchart} represents the publication search and selection procedure. This study uses five widely accepted academic databases and search engines 
    to collect peer-reviewed articles. 
    It is observed that SSL has garnered significant attention from 2018 onwards. Thus, the selection of articles is restricted from 2018. 
    Initially, 
    a total of {250} articles are selected.
    From the selected articles, the following information is logged into a registry: \textit{title, publication year, publication venue 
    , data source, objectives of the study, learning type (supervised, self-supervised, or unsupervised), contributions, limitations}, and \textit{challenges of the work}. 
    By analyzing the registry and the articles' abstract, duplicate articles and articles that do not match the scope of this survey are removed from the in-depth investigation. 
    This study reviews over 100 articles and organizes their insights for readers. However, the total citation count in the manuscript is 186, as it includes references from other sections such as the introduction, preliminary information, and benchmark datasets.



\subsection{Literature Summary}

Table~\ref{tab:researchsummary} on page \pageref{tab:researchsummary} presents a concise summary of key studies from 2016 to 2025 that explore SSL for segmentation. 

\begin{table*}[!htp]
    \centering
    \setlength{\tabcolsep}{1.2pt} 
    \renewcommand{\arraystretch}{1.1} 
    \caption{A highlight of the adaptation of self-supervised learning in different research fields. Note: \protect\seg~- Segmentation using Dice score, Recall, F1-score, or IoU, \protect\OD~- Object detection, and $\boxtimes$ - Classification.}
    \label{tab:researchsummary}
    \begin{tabular}{>{\centering\arraybackslash}m{0.6cm} 
                    >{\raggedright\arraybackslash}m{4.2cm} 
                    >{\raggedright\arraybackslash}m{3.5cm} 
                    >{\raggedright\arraybackslash}m{2.85cm} 
                    >{\raggedright\arraybackslash}m{2.85cm} 
                    >{\raggedright\arraybackslash}m{2.5cm} 
                    >{\raggedright\arraybackslash}m{1.35cm}}
    \hline
    \centering \textbf{Ref.} 
        & \centering \textbf{Methodology} 
        & \centering \textbf{Pre-processing/ Pretext} 
        & \centering \textbf{Advantages} 
        & \centering \textbf{Limitations} 
        & \centering \textbf{Dataset Used}
        & \multicolumn{1}{c}{\textbf{Results ($\%$)}} \\ \hline\hline
    
        \cite{noroozi2016unsupervised} 2016 & The backbone is pre-trained solving Jigsaw puzzle. The top layers are fine-tuned to solve a downstream task, while the backbone is frozen & Creating 9-title Jigsaw puzzle permutations on ImageNet and assigning an index to each title & Learn spatial and semantic relationships within an image without labeled data & Sensitivity to tile size \& grid configuration, Computationally expensive & D1:~PascalVOC'07 \cite{everingham2015pascal} &  $\boxtimes$ D1: 67.6 \OD~D1: 53.2 \seg~D1: 37.6 \\ \hline
        
        \cite{zhang2023self} 2017 & Layer-decomposition for generating synthetic data. A two-stage Sim2Real unsupervised self-training & Data augmentation (rotations, scaling, mirror, noise addition \& blur, change in brightness) & Fine-tuning is not required, scalability & Heavily depends on the quality and realism of the synthetic data & D1: BraTS'18~\cite{labella2024brain}, \newline D2: LiTS2017 & \seg~D1: 84.5 \seg~D2: 60.5\\ \hline

       \cite{spitzer2018improving} 2018 & Uses a self-supervised Siamese network to predict the 3D geodesic distance between image patches from histological brain sections & Extracts $1019\times1019$ patches from BigBrain 3D brain sections & Improves segmentation accuracy with limited labeled data & Relies on accurate 3D brain reconstruction & D1: BigBrain &\seg~D1: 80.0 \\ \hline

        \cite{novosel2019boosting} 2019 &  Colorization and depth prediction for semantic segmentation & Image resizing, data augmentation, depth map and grayscale generation &	Claimed that depth prediction performs better than colorization & The multi-task design can lead to increased computational & D1: Cityscapes~\cite{cordts2016cityscapes}, \newline D2: KITTI & \seg~D1: 65.0 \seg~D2: 49.0\\ \hline
    
       \cite{zeng2019sese} 2019 & SeSe-Net: It uses a dual network strategy: Worker for segmentation and Supervisor for evaluating segmentation quality &  Data augmentation with random shuffling to generate balanced datasets & It achieves performance improvements w/t $5\%$ labeled data by utilizing large unlabeled data & Due to dual-network model training can be computationally intensive & D1: Carvana, \newline D2: CMR, \newline D3: Pet & \seg~D1: 76.1 \newline \seg~D2: 67.9 \newline \seg~D3: 69.2 \\  \hline
    
        \cite{feng2019self} 2019 & Decoupling the rotation-related features from task-agnostic features & Generate multiple rotated copies of the same image ($\{0^\circ, 90^\circ, 180^\circ, 270^\circ\}$) for the rotation prediction pretext & Better generalization for downstream tasks & Increased complexity, discrete rotation angles may not capture all real-world scenarios & D1: PascalVOC'07, \newline D2: PascalVOC'12  & $\boxtimes$~D1: 74.7 \OD~D1: 58.0 \seg~D2: 45.9 \\  \hline
    
       \cite{valada2020self} 2020 & SSMA: a multimodal SSL model adaptation, dynamically recalibrates and fuses modality-specific features based on object class, spatial location, and scene context & Data augmentation (rotation, skewing, scaling, cropping, brightness \& contrast modulation, and flipping) &  Adapts feature fusion based on object class and spatial location & Requires high memory due to multi-stream architecture & D1: Cityscapes \cite{cordts2016cityscapes}, \newline D2: ScanNet, \newline D3: SYNTHIA \cite{ros2016synthia}  & \seg~D1: 82.3 \newline \seg~D2: 57.7 \newline \seg~D3: 91.3 \\ \hline
    
    
        \cite{xie2020pgl} 2020 & PGL: A local patch-based prior-guided model & Data augmentation (flipping, cropping, and scaling) & Learn meaningful local structures & Training time is high due to model complexity & D1: KiTS \cite{heller2021state}, \newline D2: BVC \cite{xu2016multi} & \seg~D1: 84.3 \seg~D2: 70.5  \\  \hline
    
    
    \cite{soliman2020s} 2020 &  An integration of supervision and self-supervision. Segmentation w/t fully-supervised fashion and reconstruction via self-supervision & An auxiliary image reconstruction  task & Achieves $18\%$ IoU improvement with only $2\%$ labeled data & Segmentation and reconstruction loss balancing is essential for optimal performance  & D1: Facies (segmented seismic images) &  \seg~D1: 62.4\\ \hline
        
        \cite{araslanov2020single} 2020 & A class aggregation function, a module for masking, and a stochastic gate work & Random rescaling, horizontal flipping, color jittering, and random crops & Produced high-quality semantic masks & Mislabelling and misleading under conditions of occlusions & D1: PascalVOC'12 & \seg~D1: 64.3 \\  \hline

    \cite{9460820} 2021 & Pre-training a backbone jointly on multimodal (image-, patch-, and pixel-level) tasks  & 3 Pretext tasks (inpainting, transformation prediction, contrastive learning) & Suitable for both high-level and low-level features learning & Multiple pretext tasks increases the complexity of the training process & D1: Potsdam, \newline D2: Vaihingen, \newline D3: Levir\_CS & \seg~D1: 70.7 \newline \seg~D2: 74.1 \newline \seg~D3: 76.5\\ \hline

    \cite{marsocci2021mare} 2021 & Use prediction of visual word distributions (OBoW) with attention for SSL and MAResU-Net for segmentation & Data augmentation (color jittering, gray-scaling, and Gaussian blurring) & Pretraining with OBoW allows the encoder to capture both global and local features & The combined SSL and attention-enhanced model require significant computational resources & D1: ISPRS Vaihingen & \seg~D1: 81.8 \\  \hline
    
        \cite{zhang2021self} 2021 &  A layer decomposition model for 3D tumor segmentation with Sim2Real synthetic data generation, and zero-shot training & Linear transformation, elastic deformation, and Gaussian blur & Disentangling tumors from the background can improve feature learning process & Computational overhead due to multi-layer processing, limited generalization & D1: BraTS'18, \newline D2: LiTS2017 & \seg~D1: 71.6 \newline \seg~D2: 40.8\\ \hline
    
       \cite{Nikita2021CVF} 2021 & SSL with momentum update to learn augmentation consistency across various augmented versions of the same image & Data augmentation (photometric noise, multi-scale fusion, and random flipping) & Achieved improvements across different backbone architectures and adaptation scenarios & Patch-size dependence, inaccurate pseudo labels can propagate errors during training & D1: Cityscapes & \seg~D1: 53.8 \\ \hline

        \cite{yang2022fully} 2022 &  A FCN patch-wise classification network trained with jigsaw puzzle proxy task for semantic segmentation & Grid-based patch generation and shuffling, Data augmentation (random mirroring and scaling) & A $5.8\%$ improvement w/t 1/6 labeled data over a baseline model trained from scratch & The patch-wise classification increases the \# of classes, complicating training & D1: PascalVOC'12 & \seg~D1: 43.4\\  \hline

      \cite{felfeliyan2023self} 2023 &  An enhanced Mask-RCNN is pre-trained with three pretext tasks via self-supervision. The model is fine-tuned using a few labeled data & Pretext tasks: localizing the distorted area, classifying the distortion type, and recovering the distorted areas & A $20\%$ increase in Dice score compared to models trained from scratch & Effectiveness depends on choice \& relevance of used distortions to the specific context & D1: Osteoarthritis Initiative (OAI) & \seg~D1: 77.0 \\  \hline

     \cite{kalapos2023self} 2023 & Self-supervised BYOL pretraining on natural and domain-specific cardiac MRI data, followed by U-Net segmentation fine-tuning for improving accuracy & Data augmentation (random resizing and cropping, horizontal flipping, brightness, and contrast) & $4-5\times$ faster fine-tuning compared to ImageNet pretraining & Less effective in extremely low-data access & D1: Adverse Conditions Dataset w/t Correspondences (ACDC) & \seg~D1: 85.0 \\  \hline

    \cite{caron2024location} 2024 & LOCA (Location-Aware) SSL. Combines patch-level clustering for dense learning and relative position prediction for spatial reasoning & Pretext (position prediction, patch-level clustering), Augmentation (flipping, cropping, color jittering, etc.) & Spatially-aware robust representation learning & Balancing position prediction \& patch-level clustering is crucial for optimal learning & D1: ADE20K~\cite{xie2021selfsupervised}, \newline D2: Cityscapes \cite{cordts2016cityscapes}, \newline D3: CamVid~\cite{brostow2009semantic}  & \seg~D1: 47.9 \newline \seg~D2: 79.8 \newline \seg~D3: 56.1 \\  \hline

    \cite{dhamale2025dual} 2025 & A dual multi-scale encoder-decoder network that combines EfficientNet B5 and attention-based pyramid vision transformer with Pixel-wise Contrastive Loss (PCL) & Boundary masks created using morphological operations, contrastive learning for pretext task & Integrated multi-scale analysis and PCL enables the model to capture contextually rich features & The PCL loss computation is complex, requires additional training time &  D1: BraTS'20, \newline D2: Kvasir-SEG, \newline D3: DFU & \seg~D1: 85.5 \newline \seg~D2: 88.1 \newline \seg~D3: 75.7\\

    \hline \hline
 
    \end{tabular}
    \end{table*}

\end{document}